\documentclass[a4paper,fleqn]{cas-sc}

\usepackage[numbers]{natbib}
\usepackage{latexsym}
\usepackage{amssymb}
\usepackage{amsmath}
\usepackage{booktabs}
\usepackage{enumitem}
\usepackage{graphicx}
\usepackage{color}
\usepackage{makecell}
\usepackage{algorithm}
\usepackage{algpseudocode}
\usepackage{xcolor}
\usepackage{colortbl}
\usepackage{makecell} 
\def\tsc#1{\csdef{#1}{\textsc{\lowercase{#1}}\xspace}}
\tsc{WGM}
\tsc{QE}

\begin{document}
\let\WriteBookmarks\relax
\def\floatpagepagefraction{1}
\def\textpagefraction{.001}
\let\printorcid\relax
\shorttitle{}

\definecolor{dark_red}{rgb}{0.5, 0, 0}
\definecolor{red}{rgb}{.9,0,0}
\newcommand{\mj}[1]{{\color{dark_red}#1}} 
\newcommand{\mjc}[1]{{\color{red}[MJ:#1]}}

\shortauthors{Jun Wang et~al.}  

\title [mode = title]{GazeCLIP: Enhancing Gaze Estimation Through Text-Guided Multimodal Learning}  

\author[1]{Jun Wang}[style=chinese]
\credit{Conceptualization, Methodology, Software, Writing - original draft preparation}
\author[1]{Hao Ruan}[style=chinese]
\credit{Conceptualization, Methodology, Software, Writing - original draft preparation}
\author[2]{Liangjian Wen}[style=chinese]
\credit{Conceptualization, Methodology, Software, Writing - original draft preparation}

\author[3]{Yong Dai}[style=chinese]
\credit{Conceptualization, Methodology, Software, Visualization, Writing - original draft preparation}
\author[4]{Mingjie Wang}[style=chinese]
\cormark[1]
\credit{Conceptualization, Methodology, Writing - Original draft preparation, Writing - review and editing, Funding acquisition}
\ead{mingjiew@zstu.edu.cn}




\affiliation[1]{organization={School of Management Science and Engineering, Southwestern University of Finance and Economics},
    city={Chengdu},
    citysep={},
    country={China}}
\affiliation[2]{organization={School of Computing and Artificial Intelligence, Southwestern University of Finance and Economics},
    city={Chengdu},
    citysep={}, 
    country={China}}

\affiliation[3]{organization={Hithink RoyalFlush Information Network Co., Ltd., China.},
    city={Hangzhou},
    citysep={}, 
    country={China}}

\affiliation[4]{organization={School of Science, Zhejiang Sci-Tech University},
    city={Hangzhou},
    citysep={}, 
    country={China}}

\cortext[cor1]{Corresponding author.}


\begin{abstract}
Visual gaze estimation, with its wide-ranging application scenarios, has garnered increasing attention within the research community. Although existing approaches infer gaze solely from image signals, recent advances in visual-language collaboration have demonstrated that the integration of linguistic information can significantly enhance performance across various visual tasks. Leveraging the remarkable transferability of large-scale Contrastive Language-Image Pre-training (CLIP) models, we address the open and urgent question of how to effectively apply linguistic cues to gaze estimation. In this work, we propose GazeCLIP, a novel gaze estimation framework that deeply explores text-face collaboration. Specifically, we introduce a meticulously designed linguistic description generator to produce text signals enriched with coarse directional cues. Furthermore, we present a CLIP-based backbone adept at characterizing text-face pairs for gaze estimation, complemented by a fine-grained multimodal fusion module that models the intricate interrelationships between heterogeneous inputs. Extensive experiments on three challenging datasets demonstrate the superiority of GazeCLIP, which achieves state-of-the-art accuracy. Our findings underscore the potential of using visual-language collaboration to advance gaze estimation and open new avenues for future research in multimodal learning for visual tasks. The implementation code and the pre-trained model will be made publicly available.
\end{abstract}





\maketitle

\section{Introduction}
In recent years, large-scale linguistic-vision models~\citep{bao2021vlmo, yu2022coca, li2021align} have emerged as transformative forces in artificial intelligence, driving significant advancements in multimodal learning. Among these, Contrastive Language-Image Pre-Training (CLIP)~\citep{radford2021learning} has garnered particular attention for its ability to bridge visual and textual modalities, achieving remarkable success across a wide range of downstream vision tasks. CLIP's architecture, built on a foundation of transformer blocks, is trained on an extensive dataset of 400 million image-text pairs, enabling it to implicitly enrich visual features with the nuanced semantics of natural language. This capability has proven invaluable in tasks such as image generation~\citep{ramesh2021zero,ramesh2022hierarchical}, visual question answering~\citep{shen2021much}, object detection~\citep{shi2022proposalclip}, semantic segmentation~\citep{xu2022simple}, and image classification~\citep{zhou2022conditional}, where the integration of linguistic guidance has consistently led to performance improvements.

However, while the benefits of linguistic-vision collaboration have been extensively explored in many visual tasks~\citep{kim2021vilt,li2022grounded,zhang2022pointclip,zhang2022can}, the field of gaze estimation, a fundamental and widely applicable visual task, remains largely untapped in this regard. Gaze estimation, which infers the direction of a person's gaze from visual data, has traditionally relied solely on image signals, overlooking the potential of leveraging linguistic information to enhance performance. This gap presents a significant opportunity, as the integration of language guidance could provide additional contextual cues, such as coarse directional information, that are inherently challenging to capture from visual data alone.

\begin{figure*}[t]
\centerline{\includegraphics[width=1.\linewidth, ,trim=0 0 0 0,clip]{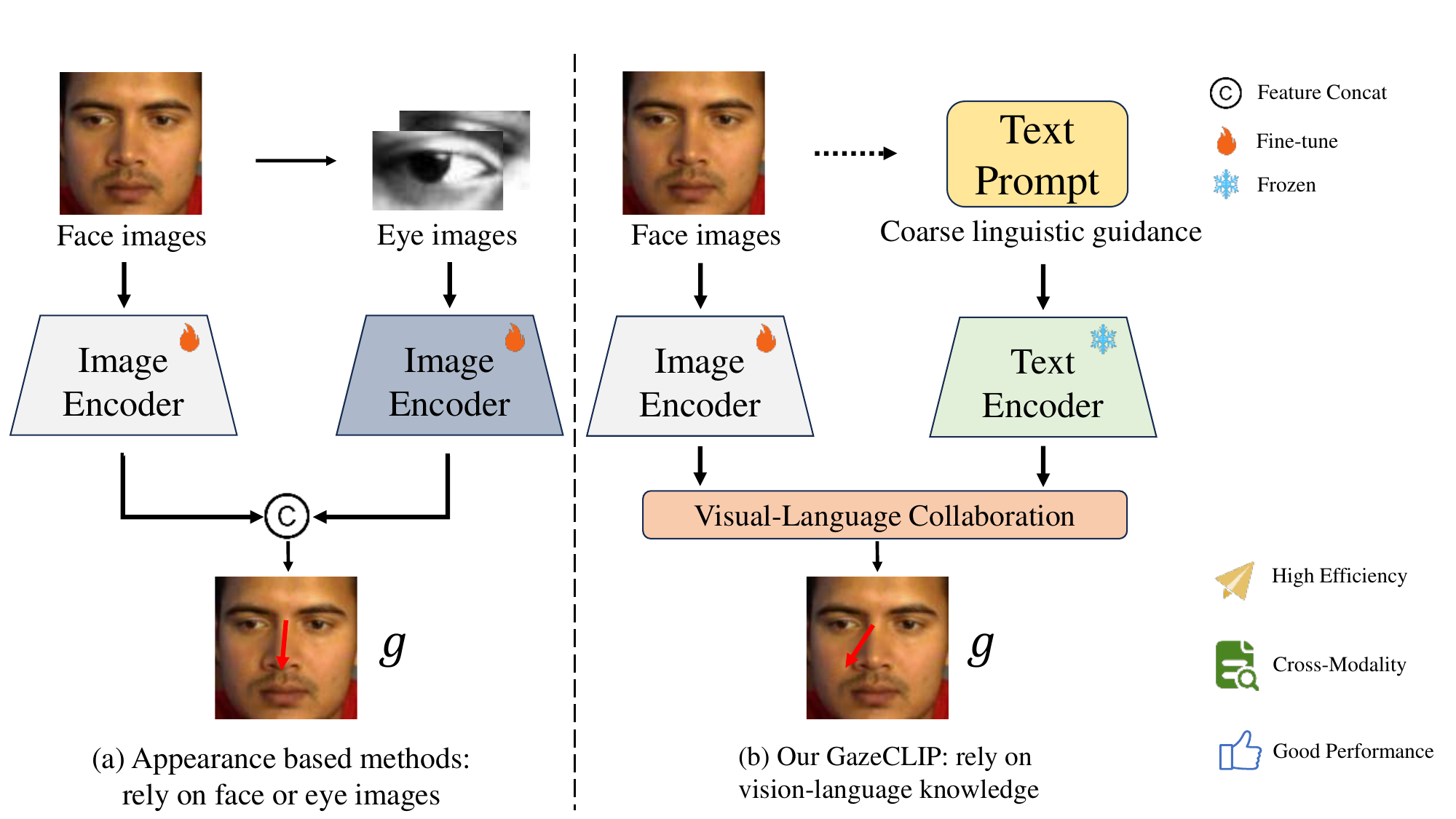}}
\caption{(a) Existing single-modal approaches directly learn gaze-oriented representations from 2D face/eye images via CNNs-based structures, whereas (b) Our proposed novel GazeCLIP delves deep into the synergistic effects of text-image features.}
\label{pic1}
\end{figure*}

Over the past decade, gaze estimation has gained increasing attention due to its wide-ranging applications, including saliency detection~\citep{wang2017deep}, virtual reality~\citep{xu2018gaze}, human-robot interaction~\citep{hempel2020slam,strazdas2022robot}, medical diagnosis~\citep{castner2020deep}, and driver fatigue estimation~\citep{yoon2019driver}. Despite its practical significance, gaze estimation faces significant challenges, such as variations in lighting, camera angles, and facial features, which can severely degrade performance. Traditional geometry-based methods~\citep{guestrin2006general} often struggle with generalization outside controlled environments, while appearance-based deep learning approaches~\citep{zhang2017mpiigaze,zhang2017s,zhang2015appearance} are limited by their reliance on single-modal facial images, making it difficult to isolate gaze-specific features from unrelated facial regions. Recent advances in high-capacity CNN models~\citep{cheng2020gaze,park2018deep,cheng2020coarse} have improved feature extraction but remain susceptible to overfitting and performance degradation under diverse conditions.

To address these challenges, we propose GazeCLIP, a novel framework that leverages the power of CLIP and linguistic guidance to advance gaze estimation. The key challenge lies in generating meaningful linguistic prompts from facial images, as traditional methods like BLIP~\citep{li2022blip} are ill-suited for this task due to their lack of gaze-specific training data. Our solution adopts a "divide-and-conquer" strategy, first identifying coarse-grained gaze directions (e.g., front, down, left, right) using a zero-shot CLIP model and then refining these directions through a fine-grained multimodal fusion module. Specifically, we pre-define text prompts such as “A photo of a face gazing [class]” and use CLIP to determine the most likely direction for each image. This approach aligns visual and textual representations through a cross-attention mechanism, enabling the model to learn nuanced gaze features while leveraging the semantic richness of natural language. Extensive experiments on three challenging datasets demonstrate the superiority of GazeCLIP, achieving state-of-the-art performance with an average reduction of 0.4°(8\% $\downarrow$) in angular error. In a nutshell, our contributions are threefold:

\begin{itemize}
    \item {\bf Novel Framework.} In this work, we introduce GazeCLIP, a novel text-guided gaze estimation framework designed to significantly enhance performance by leveraging the robust generalization capabilities of the Contrastive Language-Image Pre-training (CLIP) model across diverse downstream vision tasks. To the best of our knowledge, this represents the first endeavor to distill and harness the rich, multimodal knowledge embedded within a full-fledged pre-trained language-vision model to guide the learning process of a gaze estimation network.
    \item {\bf Fine-grained Multimodal Fusion.} We introduce a cross-attention condenser designed to finely recalibrate visual and text representations. This mechanism facilitates the nuanced alignment of image features with the semantic guidance embedded in textual signals, enhancing the learning quality of gaze features.
    \item {\bf SOTA Performance}. Extensive experiments conducted on three widely recognized datasets conclusively validate the superior performance of our proposed framework. Our innovative methodology achieves a significant performance enhancement, evidenced by an average reduction of 0.4° in angular error, which corresponds to a substantial improvement of 8\% in accuracy. By seamlessly integrating linguistic-vision collaboration with gaze estimation, GazeCLIP not only sets a new benchmark in the field but also paves the way for groundbreaking advancements in multimodal learning for visual tasks, offering a transformative perspective for future research endeavors.
\end{itemize}

\section{Related Work}

\subsection{Gaze Estimation Evolution}
Early approaches to gaze estimation primarily focused on predicting a dense array of points on a 2D screen~\citep{krafka2016eye}. However, the generalization capability of these 2D gaze estimation models is significantly hindered by the substantial variations in camera positions across different devices. To address this limitation, a series of 3D gaze estimation methods~\citep{cheng2020coarse,fischer2018rt,kellnhofer2019gaze360,zhang2017s,cheng2018appearance,chen2018appearance,wang2019generalizing} have been proposed, leveraging enriched geometric information, such as diverse shooting conditions, to infer gaze directions in real-world scenarios. In contrast to geometry-based methods, appearance-based models have gained increasing prominence within the research community due to their ability to utilize images captured by conventional cameras as input. Building on the remarkable success of deep learning techniques~\citep{he2016deep,simonyan2014very}, the performance of gaze estimation algorithms has seen substantial advancements, marking a significant leap forward in the field.

Specifically, the approach in~\citep{zhang2015appearance} pioneered the use of high-capacity convolutional neural networks (CNNs) to regress gaze directions directly from facial image inputs, marking a significant advancement in the field. Building on this, Zhang et al.~\cite{zhang2017s} introduced a spatial attention mechanism to prioritize salient facial regions within input scenes, effectively mitigating noise from irrelevant image areas. Meanwhile, Cheng et al.~\cite{cheng2018appearance} identified the inherent asymmetry between the two eyes and proposed an asymmetric regression framework comprising four distinct CNN branches. Inspired by the success of atrous convolution in image classification~\citep{yu2015multi}, Chen et al.~\cite{chen2018appearance} employed dilated convolutions to expand receptive fields without increasing computational complexity. Wang et al.~\cite{wang2019generalizing} developed a unified framework combining adversarial learning with a Bayesian approach, significantly enhancing the transferability and accuracy of gaze estimation models. The CA-Net~\citep{cheng2020coarse} further advanced the field by adopting a coarse-to-fine estimation strategy, where an initial coarse gaze direction is inferred from facial images, followed by a refinement stage using eye-specific inputs. Biswas et al.~\cite{biswas2021appearance} introduced an attention mechanism to extract critical features from eye images, while Cai et al.~\cite{cai2023source} proposed an unsupervised domain adaptation method to address cross-domain gaze estimation challenges, achieving superior performance. In~\citep{cheng2023dvgaze}, a dual-viewpoint gaze estimation network was introduced, leveraging images captured from multiple camera angles to improve robustness. 

Recent works~\citep{yin2024nerf,park2019few,cheng2022puregaze,liu2024test,liu2024uvagaze} have also focused on cross-dataset generalization, aiming to develop more universally applicable gaze estimation models. Wang et al.~\cite{wang2023investigation} demonstrated that higher-resolution images can significantly enhance performance, even with simpler backbone architectures. Despite these impressive advancements, existing gaze estimation methods have yet to exploit the rich semantic capabilities offered by modern large-scale language models (\emph{e.g.}, BERT~\citep{devlin2018bert} and CLIP~\citep{radford2021learning}). This oversight highlights a substantial opportunity for further performance improvements in gaze estimation by integrating multimodal learning paradigms.

\subsection{Language-steering Visual Models}
Recently, Large Language Models (LLMs)~\citep{radford2018improving,radford2019language,devlin2018bert} have catalyzed a paradigm shift in research, emphasizing text-oriented feature learning for a wide array of visual tasks, including crowd counting~\citep{liang2023crowdclip}, point cloud analysis~\citep{zhang2022pointclip}, and image generation~\citep{ramesh2022hierarchical}. Among these advancements, the Contrastive Language-Image Pre-training (CLIP) model~\citep{radford2021learning} has emerged as a particularly compelling framework for enhancing visual algorithms. Specifically, CLIP is designed to explore the intricate interrelationships between image and language modalities by leveraging a massive dataset of 400 million image-text pairs curated from the internet. This approach enables the model to learn rich, multimodal representations that bridge the gap between visual and textual domains, offering transformative potential for a variety of computer vision applications.

CLIP is trained using contrastive learning to maximize the cosine similarity between text and image embeddings of positive sample pairs. This innovative training paradigm eliminates the long-standing reliance on annotated image labels in the visual domain, enabling the model to learn robust multimodal representations. Remarkably, the pre-trained CLIP model achieves superior performance compared to fully supervised methods across numerous classic visual tasks, even in few-shot or zero-shot settings. Leveraging its powerful transferable capabilities, a growing body of research has explored the application of CLIP to diverse downstream tasks. Several studies~\citep{xu2022simple,hong2022avatarclip,li2022ordinalclip,liang2023crowdclip,yu2023turning} have demonstrated significant advancements in areas such as 3D avatar generation, age estimation, image aesthetics assessment, and semantic segmentation by integrating CLIP. These approaches typically employ the pre-trained CLIP model as a backbone and devise task-specific language-image interaction mechanisms. This is often achieved by predefining tailored prompts that encapsulate general linguistic descriptions of the target images, thereby enabling the model to effectively bridge visual and textual modalities for enhanced task performance.

Unfortunately, while our paper was in the process of being submitted, contemporaneous work had been published in $AAAI 2024$~\citep{yin2024clip}, which focuses on the domain generalization problem of gaze estimation by aligning CLIP image encoder primarily through distillation learning and filtering out gaze-irrelevant features through complex prompts. However, our method still has the advantage of concise prompt design, focusing on the learning and fitting ability of the model on each single dataset, and is the first to explore the introduction of multi-modal model into gaze estimation research.

\section{The proposed GazeCLIP}
This section delineates our methodology for integrating text-based knowledge into gaze estimation. We commence by elucidating the foundational principles of the CLIP model, which harnesses extensive semantic understanding of images and texts through unsupervised pre-training on a vast corpus of multimodal data. Building upon this, we provide a comprehensive exposition of our proposed framework, GazeCLIP, detailing its innovative mechanisms for leveraging linguistic guidance to enhance gaze estimation performance.

\subsection{Preliminaries on CLIP}\label{AA}
Inspired by the remarkable success of large-scale models in natural language processing~\citep{radford2018improving,radford2019language,devlin2018bert,raffel2020exploring}, recent research has increasingly focused on adapting pre-trained large models to the domain of computer vision. This paradigm shift seeks to mitigate the dependency on extensive labeled datasets for training visual tasks, as unsupervised or self-supervised learning models inherently exhibit superior transferability and generalization capabilities. By leveraging the rich, task-agnostic representations learned from vast amounts of unlabeled data, these models have demonstrated significant potential in advancing the state of the art across a wide range of visual applications.

CLIP~\citep{radford2021learning} epitomizes this approach, utilizing a vast dataset of 400 million image-text pairs curated from the web for pre-training. These text descriptions serve as natural language annotations for the corresponding images. During pre-training, batches of image-text pairs are processed through separate image and text encoders. The contrastive learning framework optimizes the cosine similarity between embeddings of matched image-text pairs while minimizing similarity for mismatched pairs. By projecting the features of both modalities into a unified embedding space, the pre-trained CLIP model demonstrates exceptional transferability, making it highly effective for a variety of downstream visual tasks, even in zero-shot scenarios. For instance, in image classification, class labels (e.g., ‘car’, ‘plane’) can be embedded into a prompt template such as “a photo of [CLASS],” consistent with CLIP's pre-training paradigm. This prompt is then encoded by the text encoder to generate class embeddings, which are compared with image embeddings to perform classification based on similarity scores. This approach underscores the versatility and robustness of CLIP in bridging visual and textual domains for diverse applications.

When adapting the CLIP model to downstream tasks, two critical considerations arise: the design of semantically aligned prompts that accurately reflect the image content, and the effective fusion of image and text embeddings to leverage their complementary information. While the original CLIP model excels in assessing image-text similarity, primarily for classification tasks, its application to diverse domains necessitates tailored modifications. In this work, we pioneer the exploration of transferring the rich, multimodal knowledge embedded in CLIP—spanning both visual and textual domains—to the field of gaze estimation. This represents the first systematic effort to harness CLIP's capabilities for this purpose, opening new avenues for advancing gaze estimation through multimodal learning.

\begin{figure*}[ht]
\centerline{\includegraphics[width=1\linewidth ,trim=0 0 10 0,clip]{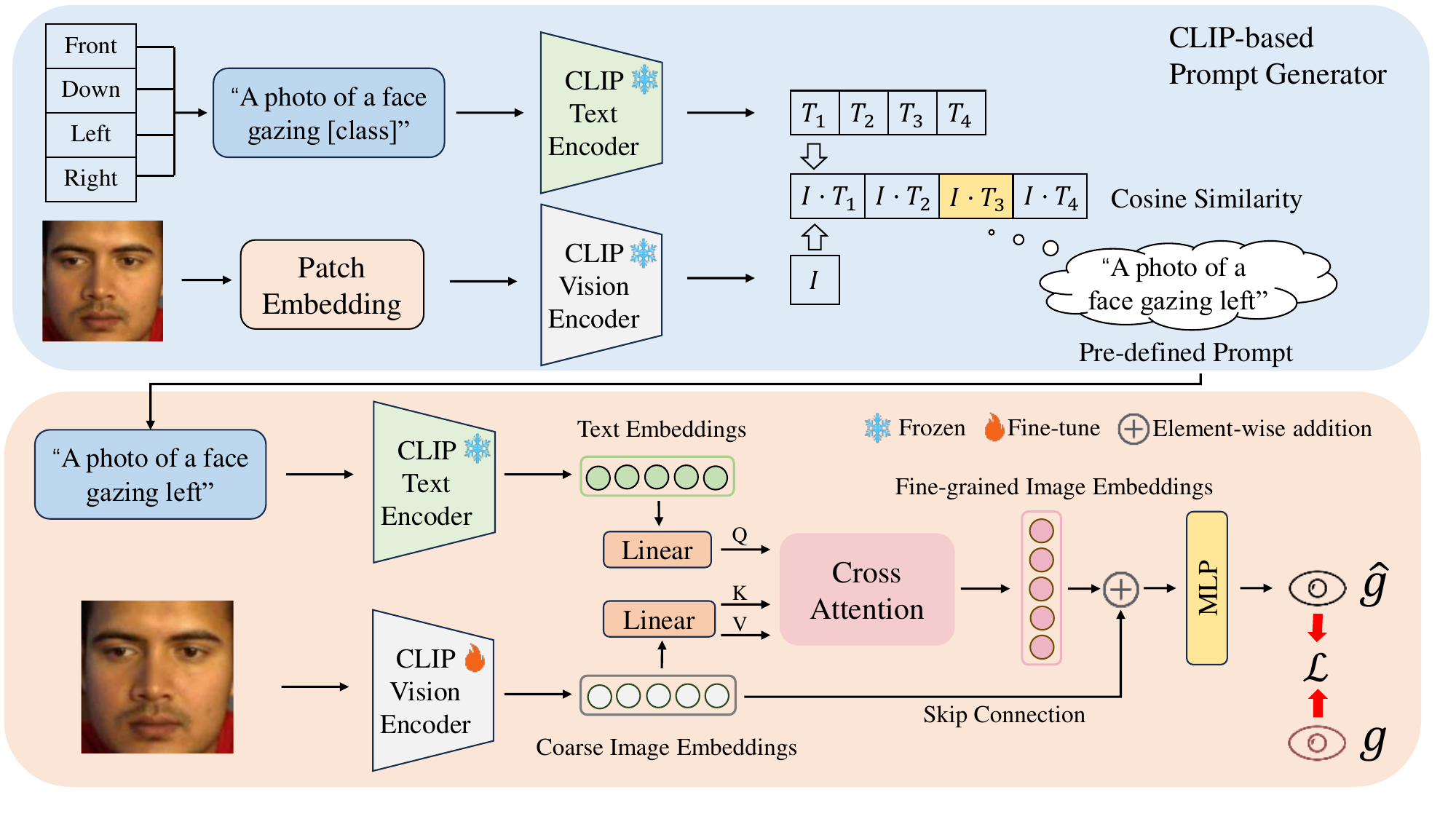}}
\caption{GazeCLIP adopts the pairs of facial images and corresponding textual description as its input and leverages the image and text encoders of the CLIP model as its foundational backbone for feature extraction. During the training phase, the image encoder is fine-tuned to adapt to the specific requirements of gaze estimation, while the parameters of the text encoder remain frozen to preserve the pre-trained linguistic knowledge. This design ensures that the model retains the robust semantic understanding of CLIP while optimizing its visual feature extraction capabilities for the task at hand.}
\label{pic2}
\end{figure*}
\subsection{The GazeCLIP Architecture}
To adapt the CLIP model for gaze estimation, we introduce GazeCLIP, as illustrated in Fig.~\ref{pic2}. The framework begins by employing the pre-trained CLIP model to generate semantically aligned prompts for each input image. Once image-text pairs are established, their respective representations are extracted using the CLIP image encoder and text encoder independently. To facilitate effective interaction between visual and textual modalities, we design a visual-linguistic interaction module based on attention mechanisms, which adaptively refines image representations by integrating linguistic guidance. These fine-grained multimodal representations are subsequently utilized for regression-based gaze prediction, enabling precise and robust estimation.

\textbf{Prompt Generation.} Ensuring the accuracy and conciseness of prompts is critical when adapting the CLIP model for downstream tasks~\citep{liang2023crowdclip,zhang2022can}. In gaze estimation, the label is a two-dimensional array representing pitch and yaw angles in three-dimensional space, making it impractical to construct prompts directly from labels, as is done in image classification tasks. To address this challenge, we pre-define the text prompt as "A photo of a face gazing [class]," where [class] represents a set of primary directions (front, down, left, right). This formulation provides a universal template for generating prompts across the gaze estimation dataset. To ensure alignment with the linguistic priors of the CLIP model, we leverage its zero-shot capability to assign a direction to each image. Specifically, we compute the cosine similarity between the image embedding and the embeddings of the predefined directional prompts, selecting the direction with the highest similarity score. This process can be viewed as a classification inference task within the CLIP framework, succinctly described as follows:
\begin{equation}
j = \arg\max_{i \in \{1, 2, 3, 4\}} \left( {cosine}(\mathbf{I}, \mathbf{T}_i) \right),\label{1}
\end{equation}
where $I$ denotes image embedding and ${T}_i$ denotes each text embedding, $i \in \{1, 2, 3, 4\}$ stands for four different directions, $cosine(\cdot)$ represents calculation of cosine similarity, $j$ means the index of direction array mentioned above with maximum similarity.

\textbf{Image Encoder.} ResNet has been extensively validated for its effectiveness across a wide range of downstream vision tasks~\citep{he2016deep}. Its residual connection mechanism enables the seamless stacking of multiple ResNet blocks, facilitating the training of deeper networks without degradation in performance. In our framework, we employ the pre-trained ResNet50 from CLIP as the image encoder, which comprises 50 ResNet blocks. The down-sampling ratio $S$ is set to 32, and the output image embedding dimension $C$ is configured to 1024. This configuration can be succinctly expressed as follows:
\begin{equation}
I=ImageEncoder(I^{\prime}),\label{2}
\end{equation}
where $I^{\prime} \in \mathbb{R}^{H \times W \times C}$ is the input image, $I$ is the image embedding with a dimension of 1024.

\textbf{Text Encoder.} In contrast to the original CLIP model and prior works that rely on generic templates such as ``a photo of a [class]'', our approach tailors the prompt to the specific requirements of the task by employing the pre-defined template ``A photo of a face gazing [class]" as input to the text encoder. This task-specific prompt is subsequently embedded into a continuous vector space. To preserve the robust linguistic priors learned during CLIP's pre-training, we freeze the parameters of the text encoder. This step can be formally defined as:
\begin{equation}
T=TextEncoder(tokenize(prompt)),\label{3}
\end{equation}
where $tokenize$ slices the complete prompts into tokens and $TextEncoder(\cdot)$ represents the standard transformer blocks~\citep{vaswani2017attention}, $T$ is the text embedding with the same dimension as the image embedding.

\textbf{Visual-linguistic Interaction Module.} To adaptively propagate coarse semantic information from linguistic features to visual features, we employ a cross-attention mechanism to model the intricate relationships between these two modalities. By focusing on the most relevant aspects of the input, the attention mechanism generates more meaningful and context-aware representations, thereby enhancing performance. This design is motivated by the strong semantic correlation between image embeddings and text embeddings derived from the pre-trained CLIP model. Prior to computing attention scores, both embeddings are projected into a shared feature space using a single linear transformation. Notably, the embedding dimension is maintained at 1024, and the number of attention heads is fixed at 1 to mitigate overfitting. We utilize the scaled dot-product attention function, which refines the attention computation by scaling the dot products between query and key vectors. This is followed by a residual connection~\citep{he2016deep} to produce fine-grained image embeddings. The entire module can be formally expressed as:
\begin{equation}
Q = linear(T), K = linear(I),\label{4}
\end{equation}
\begin{equation}
score(Q, K)=\frac{Q^T K}{\sqrt{D_k}},\label{5}
\end{equation}
\begin{equation}
V = linear(I),\label{6}
\end{equation}
\begin{equation}
\tilde I\ = Matmul(softmax(score(Q,K)), V),\label{7}
\end{equation}
\begin{equation}
\bar I\ = I + \tilde I,\label{8}
\end{equation}
where $softmax$ is applied for normalisation calculations, $D_k$ represents the dimension of embeddings and $Matmul$ represents the multiplication of two matrices.

\textbf{Regression Head.} Finally, the fine-grained image embeddings $\bar I$ are utilized for gaze prediction. A straightforward multilayer perceptron (MLP) is designed to map the 1024-dimensional image embedding into a 2-dimensional vector, representing the predicted gaze direction:
\begin{equation}
gaze(pitch, yaw) = MLP(\bar I\ ),\label{9}
\end{equation}
where the MLP contains three linear layers and two ReLu activation function layers for nonlinear transformation.

\textbf{Loss Function.} Most appearance-based gaze estimation models predict 3D gaze as gaze direction angles (yaw and pitch) in spherical coordinates. These angles are continuous values, rendering L1-loss and L2-loss suitable for optimizing the model for different datasets:
\begin{equation}
\ell_1=\frac{1}{n} \sum_{i=1}^n\left|y_i-P_i\right|,\label{10}
\end{equation}
\begin{equation}
\ell_2=\frac{1}{n} \sum_{i=1}^n\left(y_i-P_i\right)^2,\label{11}
\end{equation}
where $y_i$ is the ground truth, while $p_i$ denotes the predicted value.

\begin{algorithm}[t]
	\caption{The training process of GazeCLIP.}
	\label{a1}
\begin{algorithmic}[1]
\Statex \textbf{Input:} Image sample $\mathbf{I}^{\prime} \in \mathbb{R}^{224\times224\times3}$.
\Statex \textbf{Output:} Gaze(pitch, yaw).
\For{$epoch = 1,\ldots,E$}
    \For{$batchsize = 1,\ldots,B$}
    \State sample $\mathbf{I}^{\prime}$ from the training dataset;
    \State generate corresponding prompt through Eq.(\ref{1});  
    \State obtain image embedding ${I}$ and text embedding ${T}$ by Eq.(\ref{2}) and Eq.(\ref{3}), respectively;
    \State Calculate the attention score for each image-text pair ${I}$ and ${T}$ to get integrated embedding ${\tilde I}$;
    \State obtain fine-grained embedding $\bar I$ by Eq.(\ref{8});
    \State predict the gaze with Eq.(\ref{9});
    \State calculate the loss in (Eq.(\ref{10}),Eq.(\ref{11}));
    \State update the parameters according to $\nabla_{\theta}\mathcal{L}(\theta);$
\EndFor
\EndFor
\State \textbf{return}  model parameters $\theta$.
\end{algorithmic}
\end{algorithm}

\section{Experimental Results}
\subsection{Datasets}
To evaluate the performance of our proposed model, we conducted comprehensive experiments on three widely recognized and challenging gaze estimation datasets: MPIIFaceGaze~\citep{zhang2017mpiigaze}, EyeDiap~\citep{funes2014eyediap}, and RT-Gene~\citep{fischer2018rt}. All images in these datasets were normalized following the methodology outlined in~\citep{zhang2018revisiting} to ensure a fair comparison and guarantee that all inputs were resized uniformly before being fed into the model. In alignment with prior research, the results are reported to one decimal place. Below, we provide a detailed description of each dataset:

\textbf{MPIIFaceGaze.} MPIIGaze~\citep{zhang2015appearance} is the most commonly used benchmark dataset for unconstrained 3D gaze estimation. MPIIFaceGaze is an extension of MPIIGaze, comprising 45,000 face images captured by a laptop camera over several months, rather than just eye images from 15 subjects. Consequently, the dataset includes images with diverse backgrounds, captured at various times and under different lighting conditions. Consistent with previous works~\citep{cheng2020coarse,zhang2017s,cheng2018appearance,wang2019generalizing}, we employed the leave-one-subject-out cross-validation approach, whereby one subject's data is utilised as the test set at a time.

\textbf{EyeDiap.} The original EyeDiap dataset consists of video clips of 16 subjects. Subsequently, the images were extracted at 15-frame intervals from the video clips, in accordance with the preprocessing methodology delineated in~\citep{cheng2021appearance}. The videos collected in the screen target setting were used to obtain face images facing the screen. As videos of this type are not available for subjects No. 12 and No. 13, the processed EyeDiap dataset includes 16,000 images of 14 subjects. Since the EyeDiap dataset does not provide a standard evaluation subset, we also applied a leave-one-subject-out strategy to achieve robust results.

\textbf{RT-Gene.} The RT-Gene dataset contains 122,531 images of 15 subjects. With the aid of a series of wearable devices, the RT-Gene dataset features a greater distance between the camera and the subjects, as well as more variation in head poses and gazes compared to previous in-the-wild datasets. This is due to the use of a series of wearable devices. As a result, the dataset is more challenging. The 3-fold evaluation protocol provided by the RT-Gene dataset is followed, with the 15 subjects divided into three groups.
\subsection{Evaluation Metric}
For 3D gaze estimation, the most widely used evaluation metric is the angular gaze error, where a lower error indicates a better model. Before calculating the error, it is necessary to convert the labels and predicted results into a 3D vector, corresponding to the gaze in real 3D space. The evaluation metric is deﬁned as:
\begin{equation}
x = -cos(pitch)*sin(yaw),\label{12}
\end{equation}
\begin{equation}
y = -sin(pitch),\label{13}
\end{equation}
\begin{equation}
z = -cos(pitch)*cos(yaw),\label{14}
\end{equation}
\begin{equation}
g = (x,y,z),\label{15}
\end{equation}
\begin{equation}
\text { Angular error }=arccos(\frac{g \cdot {g^*}}{\|g\| \cdot\|{g^*}\|}),\label{16}
\end{equation}
where ${g^*}$ indicates the ground-truth and $g$ indicates the predict value. $\|\cdot\|$ represents L2-norm.
\subsection{Implementation Details}
The experiments were implemented using the PyTorch framework and conducted on an Nvidia A5000 GPU. The input images consist solely of normalized facial images with a resolution of 
224$\times$224$\times$3. For the backbone of our model, we utilized ResNet-50 (RN50)~\citep{he2016deep} for image feature extraction and a Transformer~\citep{vaswani2017attention} for text encoding. The Transformer architecture comprises 12 layers and 8 attention heads, generating features with a dimensionality of 1024, consistent with the CLIP pre-trained model. Notably, the parameters of the text encoder remain frozen throughout the training process. The regression head consists of three linear layers, outputting features with dimensions of 256, 128, and 2, respectively. The first two linear layers are followed by a ReLU activation function to introduce nonlinear transformations. The GazeCLIP model was trained on the three datasets with a batch size of 128 for 50 epochs. Optimization was performed using the Adam optimizer~\citep{kingma2014adam} with an initial learning rate of 
1e-5. The L1-loss function was applied to the MPIIFaceGaze and EyeDiap datasets, while the L2-loss function was used for the RT-Gene dataset. To dynamically adjust the learning rate during training, the MultiStepLR strategy was employed, with decay rates of 0.1 applied after the 5th and 45th epochs.


\section{Results and Analysis}

\subsection{Comparisons with the State of the Art}
We first conducted a comparative experiment to evaluate GazeCLIP against other state-of-the-art gaze estimation methodologies. Among these, only FullFace~\citep{zhang2017s} relies solely on face images as input, while other methods either directly or additionally utilize cropped eye images to enhance prediction accuracy. Furthermore, since changes in head pose are known to influence gaze estimation~\citep{zhang2015appearance}, some approaches incorporate head pose information by concatenating it with the final feature representation. For the RT-Gene dataset, which has demonstrated the effectiveness of model ensembling, we report results for both a single model and a 4-model ensemble. Notably, all existing methods have focused exclusively on image signals, primarily emphasizing the design of effective image feature extractors. For instance, Zhang et al.~\cite{zhang2017s} employ AlexNet~\citep{krizhevsky2012imagenet} to encode face images, applying spatial weights to feature maps to dynamically suppress or enhance information across different facial regions. In~\citep{fischer2018rt}, VGG-16~\citep{simonyan2014very} is used as the image encoder, with face and eye features concatenated for final prediction. Chen et al.~\cite{chen2018appearance} leverage dilated convolutions to extract high-level features, while Kellnhofer et al.~\cite{kellnhofer2019gaze360} introduce a video-based model incorporating BiLSTM for temporal modeling. In contrast, GazeCLIP introduces a novel paradigm by integrating textual guidance to refine image representations, setting it apart from these traditional approaches.

\begin{table*}
\centering
\renewcommand\arraystretch{1.5}
\caption{Quantitative Comparison of the performance on the MPIIFaceGaze, RT-Gene and EyeDiap datasets. The best results are in \textbf{bold} and the second best are \underline{underlined}.}\label{t1}
\setlength{\tabcolsep}{6mm}
\begin{tabular}{  c  c  c  c  c  }
\hline
Methods& Input & MPIIFaceGaze & RT-Gene & EyeDiap \\
\hline
FullFace~\citep{zhang2017s}  &Face         & 4.8°  & 10.0°   & 6.6°    \\
RT-Gene~\citep{fischer2018rt}  & Eyes and head pose & 4.8°         & 8.6°    & 6.4°    \\
RT-Gene(4 ensemble)~\citep{fischer2018rt}   &Eyes and head pose & 4.3°  & 7.7°    & 5.9°    \\
Dilated-Net~\citep{chen2018appearance}    &Face and eyes   & 4.8°         & 8.3°    & 6.2°    \\
Gaze360~\citep{kellnhofer2019gaze360}    &Face and eyes   & 4.1°         & -    & 5.3°    \\
CA-Net~\citep{cheng2020coarse}    &Face and eyes   & 4.1°         & 8.2°    & 5.3°    \\
AGE-Net~\citep{biswas2021appearance}    &Face and eyes   & 4.1°         & 7.4°    & -    \\
GazeTR-Hybrid~\citep{cheng2022gaze}    &Face & \underline{4.0°}     & \textbf{6.5°}   & \underline{5.2°} \\
MTGLS~\citep{ghosh2022mtgls}  &Face and eyes & 4.2°     & -   & - \\
MSGazeNet~\citep{mahmud2024multistream}    &Eyes & 4.6°     & -   & 5.8° \\
\hline
\textbf{GazeCLIP(Ours)}   &Face     & \textbf{3.5°}    & \underline{7.3}°    & \textbf{4.7°}   \\
\hline
\end{tabular}
\end{table*}

Furthermore, Cheng et al.~\cite{cheng2020coarse} explored the intrinsic correlation between face and eye features, proposing a coarse-to-fine strategy that integrates these features rather than treating them independently. Biswas et al.~\cite{biswas2021appearance} introduced an attention mechanism to enhance feature representation specifically for the eye region. Cheng et al.~\cite{cheng2022gaze} pioneered the transition from CNN-based backbones to Vision Transformers (ViT), conducting large-scale pre-training on the Eth-XGaze dataset~\citep{zhang2020eth} followed by fine-tuning on various gaze datasets. Ghosh et al.~\cite{ghosh2022mtgls} proposed a multi-task gaze representation learning framework aimed at deriving robust feature embeddings from a large corpus of non-annotated facial images. Mahmud et al.~\cite{mahmud2024multistream} introduced a pipeline that combines eye region segmentation with multi-stream gaze estimation. Notably, the inclusion of additional inputs beyond face images has consistently improved model performance, with many methods achieving comparable results. This suggests that while existing approaches have effectively tapped into high-quality visual representations, the lack of exploration into alternative information sources has limited further performance breakthroughs. Currently, the state-of-the-art performance across all three datasets is held by Gaze-TR, underscoring the potential of transformer-based architectures in advancing gaze estimation.

In contrast to the aforementioned methods, GazeCLIP also utilizes face images as input but simultaneously generates predefined prompts through a dedicated prompt generation module tailored to each image. Our primary focus lies in leveraging textual guidance to derive fine-grained image representations. As shown in Tab.~\ref{t1}, GazeCLIP achieves significant improvements of 0.5° (12\%) and 0.6° (11\%) over the previous state-of-the-art results on the MPIIFaceGaze and EyeDiap datasets, respectively. The relatively modest improvement on the RT-Gene dataset can be attributed to the fact that most images in this dataset were captured in settings where participants were positioned far from the camera. Under such conditions, the CLIP model faces challenges in making accurate coarse judgments of gaze direction compared to scenarios where images are captured using laptops. Consequently, textual guidance provides limited additional benefit in these cases. Nonetheless, the overall superior performance across all three datasets underscores the effectiveness of GazeCLIP as a robust and highly efficient network architecture for gaze estimation.

\begin{table}
\centering
\renewcommand\arraystretch{1.5}
\caption{The effect of incorporating linguistic semantic guidance through the pre-defined prompt, ''A photo of a face gazing [class]''. The results demonstrate that appropriate linguistic guidance significantly enhances model performance, highlighting the effectiveness of multimodal learning in refining gaze estimation accuracy.}\label{t2}
\setlength{\tabcolsep}{20mm}
\begin{tabular}{c c c}
\hline
Text input & Angular error & $\Delta$ \\
\hline
``Pre-defined prompt'' & \textbf{3.6°}  &-    \\
``A photo of a face''        &3.8°   &-0.2°    \\
``Empty string''        &3.9°   &-0.3°    \\
``Without text input''      & 4.4° &-0.8°   \\ 
\hline
\end{tabular}
\end{table}

\begin{table}[h]
\centering
\renewcommand\arraystretch{1.5}
\caption{The influence of freezing different encoders. Freezing the text encoder and fine-tuning the image encoder brings the best performance.}\label{t3}
\setlength{\tabcolsep}{16mm}
\begin{tabular}{ c c c }
\hline
Fixed image encoder & Fixed text encoder & Angular error \\
\hline
$\surd$ & $\surd$   &8.9°   \\
$\surd$&-   &9.0°    \\
-& $\surd$  &\textbf{3.6°}   \\ 
-& -&3.8°   \\ 
\hline
\end{tabular}
\end{table}

\subsection{Ablation Study}
To further validate the effectiveness of the individual modules within the GazeCLIP framework, we conducted three sets of ablation experiments. These experiments were performed on the MPIIFaceGaze dataset~\citep{zhang2017mpiigaze}, the most widely utilized benchmark in gaze estimation research, ensuring a comprehensive evaluation of our proposed method.

\subsubsection{Language Knowledge}
To systematically evaluate the efficacy of linguistic knowledge in GazeCLIP, we conducted a three-step ablation study. First, we retained the existing network structure but replaced the pre-defined prompts with the simplest and most intuitive description, ``A photo of a face.'' which serves as a generic representation for images in gaze estimation datasets. Next, we set the text input to an empty string, effectively removing textual guidance while keeping the network architecture intact. Finally, we eliminated the language knowledge branch entirely, fine-tuning only the image encoder for regression and removing all text-related features from the network. This step allowed us to determine whether satisfactory performance could be achieved solely through pre-trained image encoders, independent of linguistic information. Through this progressive analysis, we aimed to isolate and quantify the contribution of language knowledge to the overall performance of GazeCLIP.

As summarized in Tab.~\ref{t2}, the results demonstrate that only using the original CLIP image encoder, as in previous research architectures that rely solely on visual features, does not yield satisfactory performance. This is likely due to the fact that a more complex backbone often leads to a more severe overfitting problem in gaze estimation, especially when the size of the dataset is relatively small in comparison to the pre-trained dataset of CLIP. And it is a interesting result that when the original network structure was maintained but the defined prompt was replaced with empty strings, the performance was significantly enhanced by 0.5° (4.4° → 3.9°). But when the prompt becomes a more reasonable language expression, the error is reduced further. Instead of taking the same simple expression, using a pre-defined prompt, "A photo of a face gazing [class]" for a general language expression of the gaze estimation images with rough directions as the input text achieved the best results. These findings demonstrate the significance of text features and the effectiveness of appropriately designed prompts in our framework.

\begin{figure}[h!]
\centerline{\includegraphics[width=0.8\linewidth, trim=310 150 300 150,clip]{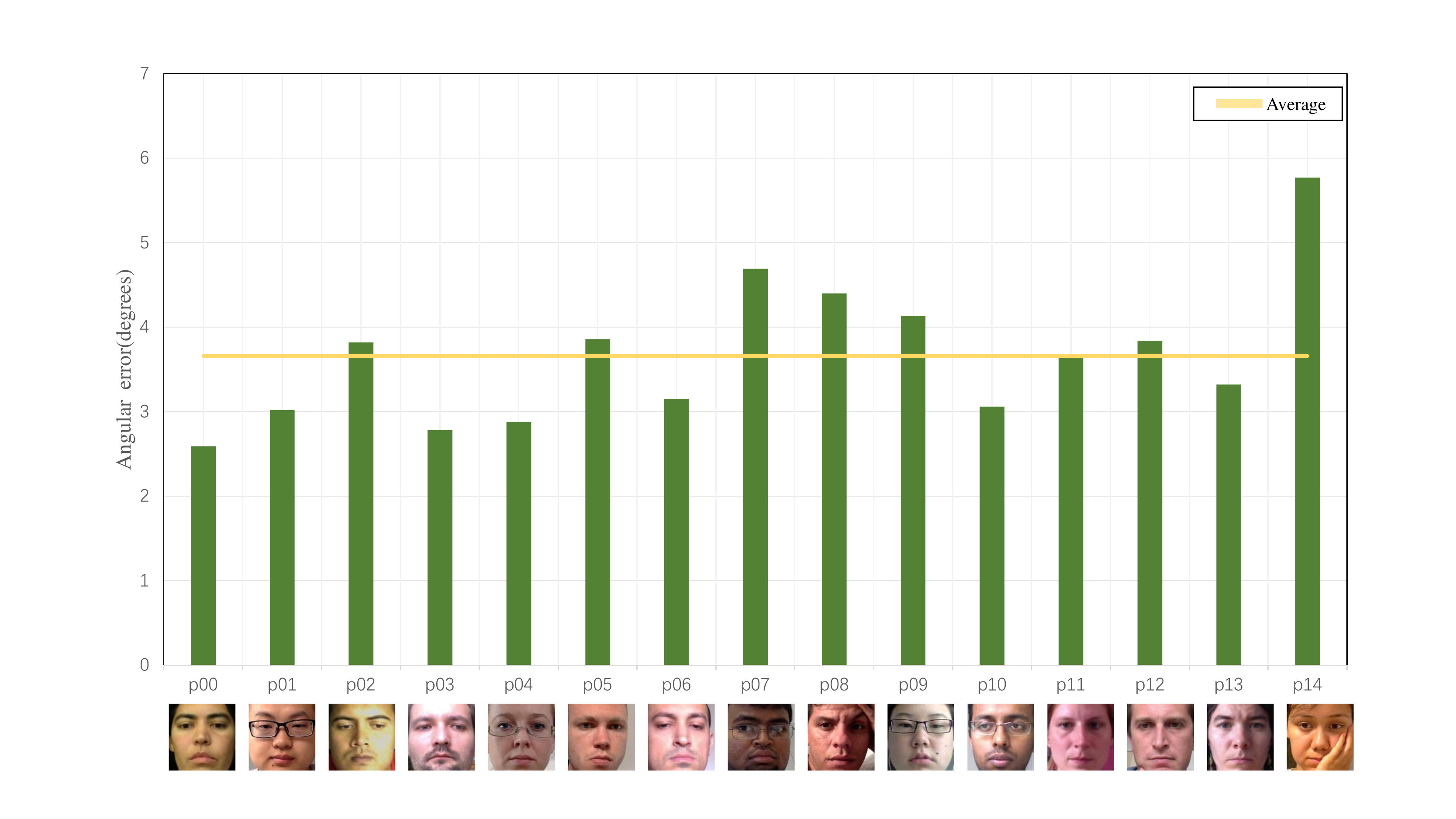}}
\caption{The results of an ablation study evaluating different feature fusion approaches. The visual-linguistic interaction module, which leverages a cross-attention mechanism combined with residual connections, demonstrates superior capability in effectively integrating features from both visual and textual modalities, leading to enhanced performance in gaze estimation.}
\label{pic8}
\end{figure}

\begin{figure}[h!]
\centerline{\includegraphics[width=0.9\linewidth,, trim=30 0 80 0,clip]{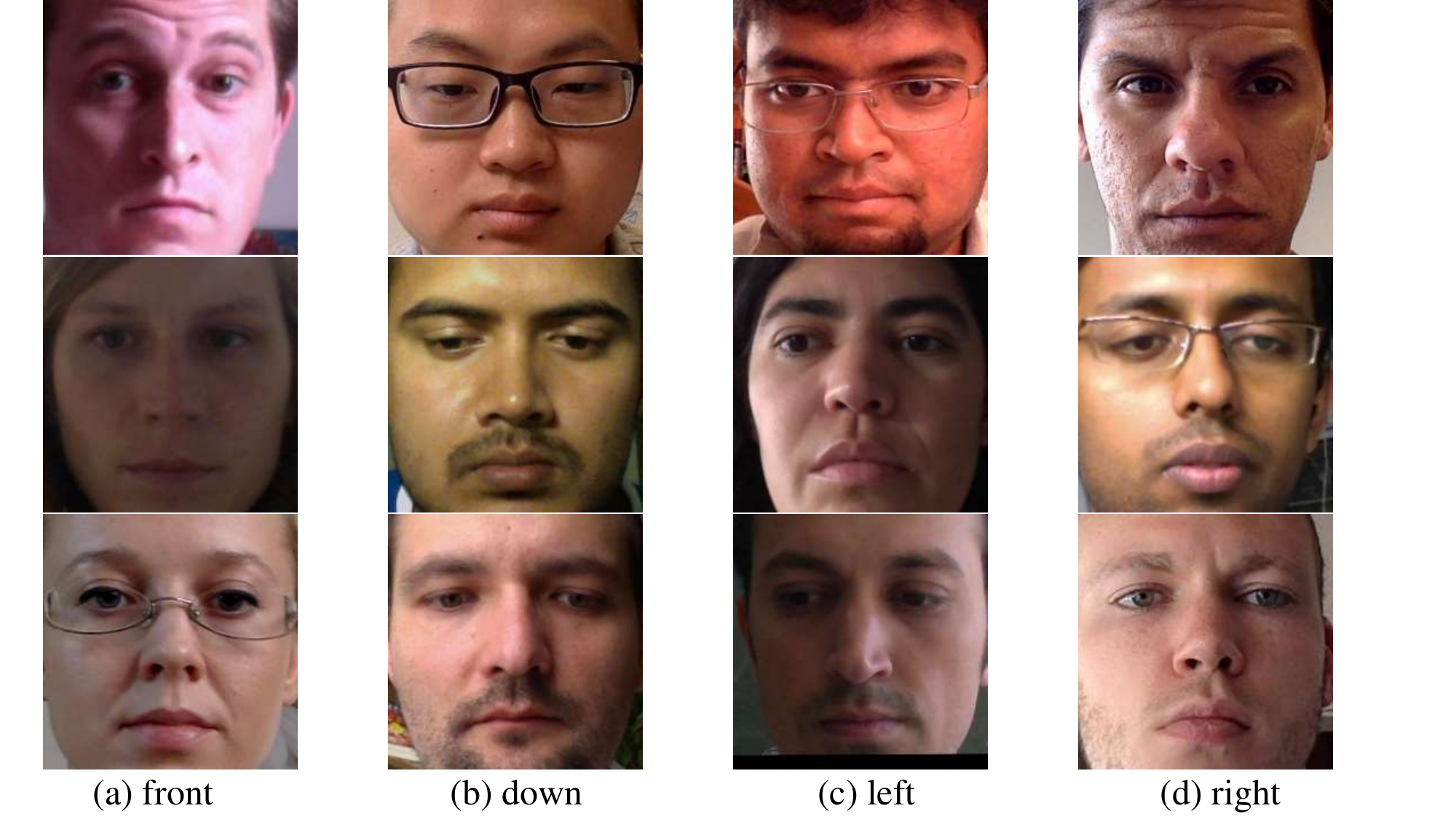}}
\caption{Images assigned in different coarse directions including fornt, down, left and right.}
\label{pic4}
\end{figure}

\begin{figure}[h!]
\centerline{\includegraphics[width=1\linewidth, trim=0 150 0 0,clip]{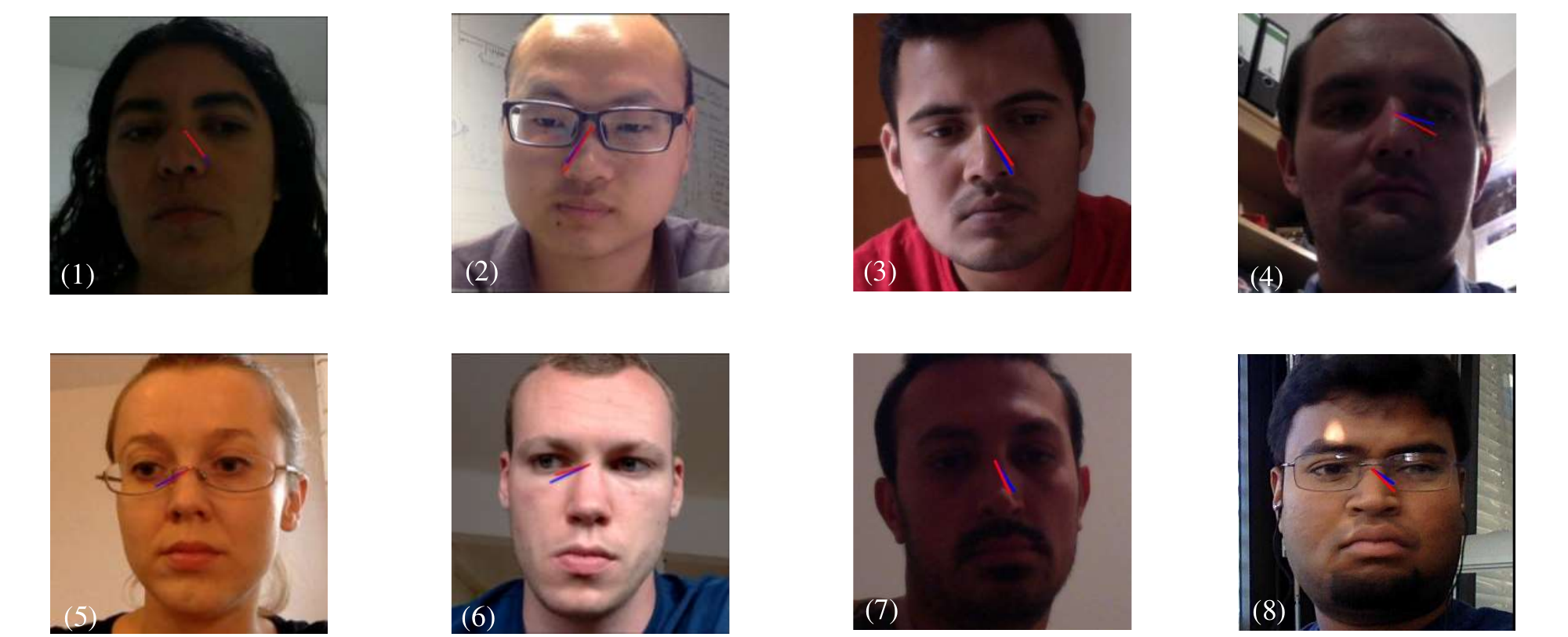}}
\caption{Visualization of inferred results.  Red lines represent ground-truth annotations, while blue lines indicate model predictions.}
\label{pic5}
\end{figure}

\subsubsection{Fixing Different Encoders}
Furthermore, the impact of freezing the image encoder and text encoder was investigated. By selectively freezing and unfreezing these encoders, four distinct configurations were evaluated. As shown in Tab.~\ref{t3}, when the original CLIP model is used directly with the visual-linguistic interaction module and a regression head (i.e., both the image and text encoders remain frozen and their parameters are not optimized), the angular error is 8.9°. Unfreezing the text encoder did not lead to a significant improvement in performance, with the angular error increasing slightly from 8.9° to 9.0°. However, when the image encoder is fine-tuned, a substantial improvement is observed, with the error decreasing from 8.9° to 3.8°. Interestingly, when both the image and text encoders are fine-tuned, the angular error increases marginally from 3.6° to 3.8°. These results suggest that, regardless of whether the image encoder is frozen, configurations with a fixed text encoder consistently deliver better performance.
\begin{figure}[h!]
\centerline{\includegraphics[width=0.8\linewidth, trim=200 300 200 50,clip]{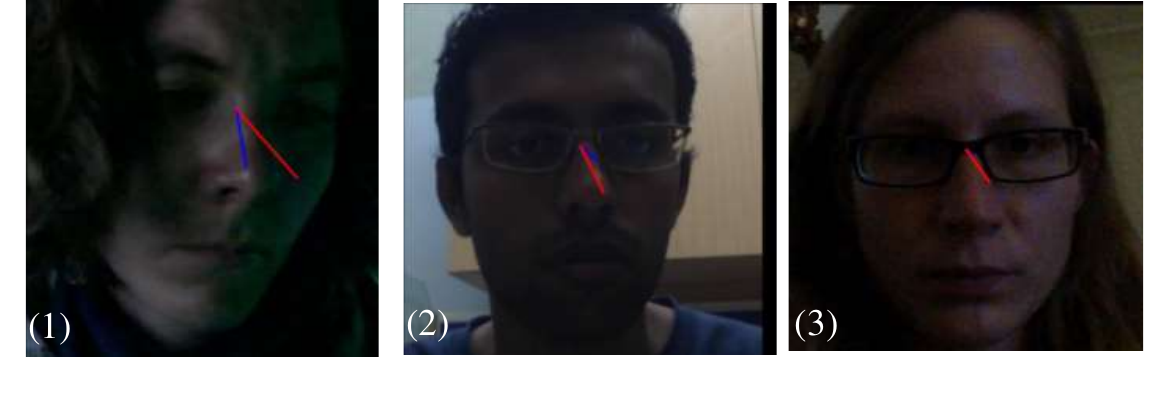}}
\caption{Failure cases observed in low-luminosity conditions. Without additional eye-specific hints or specialized focus on the eye region, the model exhibits degraded performance.}
\label{pic6}
\end{figure}
It is hypothesized that this phenomenon arises because the pre-trained CLIP model already encapsulates rich language priors that enable effective natural language embeddings, and fine-tuning the text encoder may disrupt these learned representations. Additionally, since the CLIP model is pre-trained on a diverse range of internet images, certain characteristics specific to gaze estimation datasets may be underrepresented. As noted in~\citep{radford2021learning}, the CLIP model demonstrates strong zero-shot performance across many image classification datasets. However, it struggles with simpler datasets such as MNIST~\citep{lecun1998gradient}, likely because the pre-training data for CLIP lacks examples of handwritten digits like those in MNIST. This underscores the importance of fine-tuning the image encoder to adapt the model to specific tasks, particularly when the target dataset contains features that are underrepresented in the pre-trained data.
\begin{table}
\centering
\renewcommand\arraystretch{1.5}
\caption{Results obtained using different backbones. ResNet, with fewer model parameters, demonstrates more stable performance on small-scale datasets.}\label{t4}
\setlength{\tabcolsep}{14mm}
\begin{tabular}{c c c c}
\hline
Backbone & Params    & Angular error & $\Delta$ \\
\hline
RN50     &105.810(M) & \textbf{3.5°}  &-    \\
RN101    &120.869(M) &3.6°   &-0.1°    \\
ViT/B-32 &152.458(M) &3.7°   &-0.2°    \\
\hline
\end{tabular}
\end{table}

\subsubsection{Feature Fusion Method}
The next phase of the investigation focused on evaluating the impact of various feature fusion techniques during the training process. In addition to the previously discussed visual-linguistic interaction module, two other commonly used methods were explored: concatenation and summation. Specifically, the image and text embeddings derived from the CLIP encoders were either directly concatenated or added together. Since the dimensions of the two features are identical, concatenation results in a doubling of the feature dimension. This, in turn, requires an adjustment to the input dimension of the first linear layer in the regression head. On the other hand, the summation method and the cross-attention mechanism with residual connection in the visual-linguistic interaction module preserve the original feature dimension without any changes.

The results, as illustrated in Fig.~\ref{pic8}, reveal that the concatenation and addition methods produce angular errors of 4.2° and 4.4°, respectively. In comparison, the cross-attention mechanism with residual connection achieves a lower angular error of 3.6°, outperforming the other two methods by reducing the error by 0.6° and 0.8°. These findings underscore the effectiveness of the visual-linguistic interaction module in GazeCLIP, which successfully bridges the gap between linguistic and visual knowledge extracted from the CLIP model, thereby enhancing performance in gaze estimation tasks.

\subsubsection{Parameters Comparison}
Additionally, we expand the parameter size of GazeCLIP by replacing the image backbone with larger architectures, specifically ResNet101 and ViT/B-32. However, further increasing the number of parameters or switching to an even more complex backbone, such as ViT/B-16, led to out-of-memory issues. As shown in Tab.~\ref{t4}, the increase in parameters and the correspondingly more complex image feature representation processes did not effectively reduce prediction errors. This observation underscores the tendency for overfitting in gaze estimation tasks. Additionally, it is worth noting that vision transformers (ViTs) typically require significantly more training data compared to convolutional neural networks (CNNs) to fully realize their advantages~\cite{dosovitskiy2020image}. These findings highlight the challenges of scaling up model complexity for gaze estimation and suggest the need for careful consideration of architecture design and dataset size.


\subsection{Visual Illustration}
In this section, we begin by presenting several examples of coarse gaze directions corresponding to images, as generated by the prompt module in Fig.~\ref{pic4}. Notably, the most frequently occurring prompt for the majority of images is ``A photo of a face gazing front'' as participants typically position themselves directly in front of the camera. For other gaze directions, CLIP’s zero-shot capability occasionally fails to produce judgments that align fully with human visual perception, sometimes even yielding opposite results. Nevertheless, to provide effective guidance for the image embeddings during the cross-attention stage, it is essential to generate prompts that closely align with the linguistic priors embedded in the CLIP model. As such, it is acceptable that the generated descriptions may not always match our intuitive expectations. Next, we showcase visual results of our method in Fig.~\ref{pic5}. These images demonstrate the effectiveness of our approach across a variety of scenarios, including variations in gender, facial appearance, lighting conditions, and the presence or absence of glasses. The results highlight the robustness and adaptability of our method in diverse real-world settings.

\subsection{Failure Cases and Discussion}
 Although GazeCLIP demonstrates strong performance on most images, we also present and analyze cases where the results are less satisfactory. As shown in Fig.~\ref{pic6}, these instances typically involve poor lighting conditions, leading to blurred facial features. Additionally, in many of these samples, the eye area is partially obscured by hair, glasses, or other factors, making it more challenging for the model to extract meaningful eye features. Even human observers would find it difficult to determine the true gaze direction in such scenarios. Nevertheless, with the aid of language guidance, the model is still able to provide a rough estimate of the gaze direction that aligns somewhat with the ground truth label. We attribute the suboptimal performance in these cases to the fact that our method does not explicitly leverage additional eye-specific images or incorporate focused attention mechanisms on the eye region. Addressing this limitation by exploring ways to enhance the model’s ability to prioritize and analyze eye features remains an important direction for future research.

\section{Conclusion}
In this paper, we introduce GazeCLIP, a novel framework for accurate gaze estimation based on the vision-language pre-trained model, CLIP. Our approach leverages language knowledge by first establishing consistent linguistic expressions for all images and then fusing visual and textual features to enhance the effectiveness of gaze estimation. The initial phase involved designing a module specifically tailored to generate prompts using the CLIP model. A key aspect of our method is the establishment of a robust connection between visual and textual features. To achieve this, we developed a visual-linguistic interaction module aimed at enriching image representations for improved gaze prediction. Extensive experimental results demonstrate that GazeCLIP achieves strong performance on three widely recognized publicly available datasets. Additionally, we conducted experiments to evaluate the contributions of different modules within our framework. We believe this work offers valuable insights and paves the way for future research in gaze estimation by integrating visual-language knowledge. And in future work, we are ready to explore the deep understanding of large multi-modal models for text and images to help achieve more accurate and robust gaze estimation.


\bibliographystyle{cas-model2-names}


\bibliography{cas-dc-template}

\begin{thebibliography}{67}
\expandafter\ifx\csname natexlab\endcsname\relax\def\natexlab#1{#1}\fi
\providecommand{\url}[1]{\texttt{#1}}
\providecommand{\href}[2]{#2}
\providecommand{\path}[1]{#1}
\providecommand{\DOIprefix}{doi:}
\providecommand{\ArXivprefix}{arXiv:}
\providecommand{\URLprefix}{URL: }
\providecommand{\Pubmedprefix}{pmid:}
\providecommand{\doi}[1]{\href{http://dx.doi.org/#1}{\path{#1}}}
\providecommand{\Pubmed}[1]{\href{pmid:#1}{\path{#1}}}
\providecommand{\bibinfo}[2]{#2}
\ifx\xfnm\relax \def\xfnm[#1]{\unskip,\space#1}\fi
\bibitem[{Bao et~al.(2021)Bao, Wang, Dong, Liu, Mohammed, Aggarwal, Som and Wei}]{bao2021vlmo}
\bibinfo{author}{Bao, H.}, \bibinfo{author}{Wang, W.}, \bibinfo{author}{Dong, L.}, \bibinfo{author}{Liu, Q.}, \bibinfo{author}{Mohammed, O.K.}, \bibinfo{author}{Aggarwal, K.}, \bibinfo{author}{Som, S.}, \bibinfo{author}{Wei, F.}, \bibinfo{year}{2021}.
\newblock \bibinfo{title}{Vlmo: Unified vision-language pre-training with mixture-of-modality-experts}.
\newblock \bibinfo{journal}{arXiv preprint arXiv:2111.02358} .
\bibitem[{Biswas et~al.(2021)}]{biswas2021appearance}
\bibinfo{author}{Biswas, P.}, et~al., \bibinfo{year}{2021}.
\newblock \bibinfo{title}{Appearance-based gaze estimation using attention and difference mechanism}, in: \bibinfo{booktitle}{Proceedings of the IEEE/CVF conference on computer vision and pattern recognition}, pp. \bibinfo{pages}{3143--3152}.
\bibitem[{Cai et~al.(2023)Cai, Zeng, Shan and Chen}]{cai2023source}
\bibinfo{author}{Cai, X.}, \bibinfo{author}{Zeng, J.}, \bibinfo{author}{Shan, S.}, \bibinfo{author}{Chen, X.}, \bibinfo{year}{2023}.
\newblock \bibinfo{title}{Source-free adaptive gaze estimation by uncertainty reduction}, in: \bibinfo{booktitle}{Proceedings of the IEEE/CVF Conference on Computer Vision and Pattern Recognition}, pp. \bibinfo{pages}{22035--22045}.
\bibitem[{Castner et~al.(2020)Castner, Kuebler, Scheiter, Richter, Eder, H{\"u}ttig, Keutel and Kasneci}]{castner2020deep}
\bibinfo{author}{Castner, N.}, \bibinfo{author}{Kuebler, T.C.}, \bibinfo{author}{Scheiter, K.}, \bibinfo{author}{Richter, J.}, \bibinfo{author}{Eder, T.}, \bibinfo{author}{H{\"u}ttig, F.}, \bibinfo{author}{Keutel, C.}, \bibinfo{author}{Kasneci, E.}, \bibinfo{year}{2020}.
\newblock \bibinfo{title}{Deep semantic gaze embedding and scanpath comparison for expertise classification during opt viewing}, in: \bibinfo{booktitle}{ACM symposium on eye tracking research and applications}, pp. \bibinfo{pages}{1--10}.
\bibitem[{Chen and Shi(2018)}]{chen2018appearance}
\bibinfo{author}{Chen, Z.}, \bibinfo{author}{Shi, B.E.}, \bibinfo{year}{2018}.
\newblock \bibinfo{title}{Appearance-based gaze estimation using dilated-convolutions}, in: \bibinfo{booktitle}{Asian Conference on Computer Vision}, \bibinfo{organization}{Springer}. pp. \bibinfo{pages}{309--324}.
\bibitem[{Cheng et~al.(2022)Cheng, Bao and Lu}]{cheng2022puregaze}
\bibinfo{author}{Cheng, Y.}, \bibinfo{author}{Bao, Y.}, \bibinfo{author}{Lu, F.}, \bibinfo{year}{2022}.
\newblock \bibinfo{title}{Puregaze: Purifying gaze feature for generalizable gaze estimation}, in: \bibinfo{booktitle}{Proceedings of the AAAI Conference on Artificial Intelligence}, pp. \bibinfo{pages}{436--443}.
\bibitem[{Cheng et~al.(2020a)Cheng, Huang, Wang, Qian and Lu}]{cheng2020coarse}
\bibinfo{author}{Cheng, Y.}, \bibinfo{author}{Huang, S.}, \bibinfo{author}{Wang, F.}, \bibinfo{author}{Qian, C.}, \bibinfo{author}{Lu, F.}, \bibinfo{year}{2020}a.
\newblock \bibinfo{title}{A coarse-to-fine adaptive network for appearance-based gaze estimation}, in: \bibinfo{booktitle}{Proceedings of the AAAI Conference on Artificial Intelligence}, pp. \bibinfo{pages}{10623--10630}.
\bibitem[{Cheng and Lu(2022)}]{cheng2022gaze}
\bibinfo{author}{Cheng, Y.}, \bibinfo{author}{Lu, F.}, \bibinfo{year}{2022}.
\newblock \bibinfo{title}{Gaze estimation using transformer}, in: \bibinfo{booktitle}{2022 26th International Conference on Pattern Recognition (ICPR)}, \bibinfo{organization}{IEEE}. pp. \bibinfo{pages}{3341--3347}.
\bibitem[{Cheng and Lu(2023)}]{cheng2023dvgaze}
\bibinfo{author}{Cheng, Y.}, \bibinfo{author}{Lu, F.}, \bibinfo{year}{2023}.
\newblock \bibinfo{title}{Dvgaze: Dual-view gaze estimation}, in: \bibinfo{booktitle}{Proceedings of the IEEE/CVF International Conference on Computer Vision}, pp. \bibinfo{pages}{20632--20641}.
\bibitem[{Cheng et~al.(2018)Cheng, Lu and Zhang}]{cheng2018appearance}
\bibinfo{author}{Cheng, Y.}, \bibinfo{author}{Lu, F.}, \bibinfo{author}{Zhang, X.}, \bibinfo{year}{2018}.
\newblock \bibinfo{title}{Appearance-based gaze estimation via evaluation-guided asymmetric regression}, in: \bibinfo{booktitle}{Proceedings of the European conference on computer vision (ECCV)}, pp. \bibinfo{pages}{100--115}.
\bibitem[{Cheng et~al.(2021)Cheng, Wang, Bao and Lu}]{cheng2021appearance}
\bibinfo{author}{Cheng, Y.}, \bibinfo{author}{Wang, H.}, \bibinfo{author}{Bao, Y.}, \bibinfo{author}{Lu, F.}, \bibinfo{year}{2021}.
\newblock \bibinfo{title}{Appearance-based gaze estimation with deep learning: A review and benchmark}.
\newblock \bibinfo{journal}{arXiv preprint arXiv:2104.12668} .
\bibitem[{Cheng et~al.(2020b)Cheng, Zhang, Lu and Sato}]{cheng2020gaze}
\bibinfo{author}{Cheng, Y.}, \bibinfo{author}{Zhang, X.}, \bibinfo{author}{Lu, F.}, \bibinfo{author}{Sato, Y.}, \bibinfo{year}{2020}b.
\newblock \bibinfo{title}{Gaze estimation by exploring two-eye asymmetry}.
\newblock \bibinfo{journal}{IEEE Transactions on Image Processing} \bibinfo{volume}{29}, \bibinfo{pages}{5259--5272}.
\bibitem[{Devlin et~al.(2018)Devlin, Chang, Lee and Toutanova}]{devlin2018bert}
\bibinfo{author}{Devlin, J.}, \bibinfo{author}{Chang, M.W.}, \bibinfo{author}{Lee, K.}, \bibinfo{author}{Toutanova, K.}, \bibinfo{year}{2018}.
\newblock \bibinfo{title}{Bert: Pre-training of deep bidirectional transformers for language understanding}.
\newblock \bibinfo{journal}{arXiv preprint arXiv:1810.04805} .
\bibitem[{Dosovitskiy et~al.(2020)Dosovitskiy, Beyer, Kolesnikov, Weissenborn, Zhai, Unterthiner, Dehghani, Minderer, Heigold, Gelly et~al.}]{dosovitskiy2020image}
\bibinfo{author}{Dosovitskiy, A.}, \bibinfo{author}{Beyer, L.}, \bibinfo{author}{Kolesnikov, A.}, \bibinfo{author}{Weissenborn, D.}, \bibinfo{author}{Zhai, X.}, \bibinfo{author}{Unterthiner, T.}, \bibinfo{author}{Dehghani, M.}, \bibinfo{author}{Minderer, M.}, \bibinfo{author}{Heigold, G.}, \bibinfo{author}{Gelly, S.}, et~al., \bibinfo{year}{2020}.
\newblock \bibinfo{title}{An image is worth 16x16 words: Transformers for image recognition at scale}.
\newblock \bibinfo{journal}{arXiv preprint arXiv:2010.11929} .
\bibitem[{Fischer et~al.(2018)Fischer, Chang and Demiris}]{fischer2018rt}
\bibinfo{author}{Fischer, T.}, \bibinfo{author}{Chang, H.J.}, \bibinfo{author}{Demiris, Y.}, \bibinfo{year}{2018}.
\newblock \bibinfo{title}{Rt-gene: Real-time eye gaze estimation in natural environments}, in: \bibinfo{booktitle}{Proceedings of the European conference on computer vision (ECCV)}, pp. \bibinfo{pages}{334--352}.
\bibitem[{Funes~Mora et~al.(2014)Funes~Mora, Monay and Odobez}]{funes2014eyediap}
\bibinfo{author}{Funes~Mora, K.A.}, \bibinfo{author}{Monay, F.}, \bibinfo{author}{Odobez, J.M.}, \bibinfo{year}{2014}.
\newblock \bibinfo{title}{Eyediap: A database for the development and evaluation of gaze estimation algorithms from rgb and rgb-d cameras}, in: \bibinfo{booktitle}{Proceedings of the symposium on eye tracking research and applications}, pp. \bibinfo{pages}{255--258}.
\bibitem[{Ghosh et~al.(2022)Ghosh, Hayat, Dhall and Knibbe}]{ghosh2022mtgls}
\bibinfo{author}{Ghosh, S.}, \bibinfo{author}{Hayat, M.}, \bibinfo{author}{Dhall, A.}, \bibinfo{author}{Knibbe, J.}, \bibinfo{year}{2022}.
\newblock \bibinfo{title}{Mtgls: Multi-task gaze estimation with limited supervision}, in: \bibinfo{booktitle}{Proceedings of the IEEE/CVF winter conference on applications of computer vision}, pp. \bibinfo{pages}{3223--3234}.
\bibitem[{Guestrin and Eizenman(2006)}]{guestrin2006general}
\bibinfo{author}{Guestrin, E.D.}, \bibinfo{author}{Eizenman, M.}, \bibinfo{year}{2006}.
\newblock \bibinfo{title}{General theory of remote gaze estimation using the pupil center and corneal reflections}.
\newblock \bibinfo{journal}{IEEE Transactions on biomedical engineering} \bibinfo{volume}{53}, \bibinfo{pages}{1124--1133}.
\bibitem[{He et~al.(2016)He, Zhang, Ren and Sun}]{he2016deep}
\bibinfo{author}{He, K.}, \bibinfo{author}{Zhang, X.}, \bibinfo{author}{Ren, S.}, \bibinfo{author}{Sun, J.}, \bibinfo{year}{2016}.
\newblock \bibinfo{title}{Deep residual learning for image recognition}, in: \bibinfo{booktitle}{Proceedings of the IEEE conference on computer vision and pattern recognition}, pp. \bibinfo{pages}{770--778}.
\bibitem[{Hempel and Al-Hamadi(2020)}]{hempel2020slam}
\bibinfo{author}{Hempel, T.}, \bibinfo{author}{Al-Hamadi, A.}, \bibinfo{year}{2020}.
\newblock \bibinfo{title}{Slam-based multistate tracking system for mobile human-robot interaction}, in: \bibinfo{booktitle}{International Conference on Image Analysis and Recognition}, \bibinfo{organization}{Springer}. pp. \bibinfo{pages}{368--376}.
\bibitem[{Hong et~al.(2022)Hong, Zhang, Pan, Cai, Yang and Liu}]{hong2022avatarclip}
\bibinfo{author}{Hong, F.}, \bibinfo{author}{Zhang, M.}, \bibinfo{author}{Pan, L.}, \bibinfo{author}{Cai, Z.}, \bibinfo{author}{Yang, L.}, \bibinfo{author}{Liu, Z.}, \bibinfo{year}{2022}.
\newblock \bibinfo{title}{Avatarclip: Zero-shot text-driven generation and animation of 3d avatars}.
\newblock \bibinfo{journal}{arXiv preprint arXiv:2205.08535} .
\bibitem[{Kellnhofer et~al.(2019)Kellnhofer, Recasens, Stent, Matusik and Torralba}]{kellnhofer2019gaze360}
\bibinfo{author}{Kellnhofer, P.}, \bibinfo{author}{Recasens, A.}, \bibinfo{author}{Stent, S.}, \bibinfo{author}{Matusik, W.}, \bibinfo{author}{Torralba, A.}, \bibinfo{year}{2019}.
\newblock \bibinfo{title}{Gaze360: Physically unconstrained gaze estimation in the wild}, in: \bibinfo{booktitle}{Proceedings of the IEEE/CVF international conference on computer vision}, pp. \bibinfo{pages}{6912--6921}.
\bibitem[{Kim et~al.(2021)Kim, Son and Kim}]{kim2021vilt}
\bibinfo{author}{Kim, W.}, \bibinfo{author}{Son, B.}, \bibinfo{author}{Kim, I.}, \bibinfo{year}{2021}.
\newblock \bibinfo{title}{Vilt: Vision-and-language transformer without convolution or region supervision}, in: \bibinfo{booktitle}{International conference on machine learning}, \bibinfo{organization}{PMLR}. pp. \bibinfo{pages}{5583--5594}.
\bibitem[{Kingma and Ba(2014)}]{kingma2014adam}
\bibinfo{author}{Kingma, D.P.}, \bibinfo{author}{Ba, J.}, \bibinfo{year}{2014}.
\newblock \bibinfo{title}{Adam: A method for stochastic optimization}.
\newblock \bibinfo{journal}{arXiv preprint arXiv:1412.6980} .
\bibitem[{Krafka et~al.(2016)Krafka, Khosla, Kellnhofer, Kannan, Bhandarkar, Matusik and Torralba}]{krafka2016eye}
\bibinfo{author}{Krafka, K.}, \bibinfo{author}{Khosla, A.}, \bibinfo{author}{Kellnhofer, P.}, \bibinfo{author}{Kannan, H.}, \bibinfo{author}{Bhandarkar, S.}, \bibinfo{author}{Matusik, W.}, \bibinfo{author}{Torralba, A.}, \bibinfo{year}{2016}.
\newblock \bibinfo{title}{Eye tracking for everyone}, in: \bibinfo{booktitle}{Proceedings of the IEEE conference on computer vision and pattern recognition}, pp. \bibinfo{pages}{2176--2184}.
\bibitem[{Krizhevsky et~al.(2012)Krizhevsky, Sutskever and Hinton}]{krizhevsky2012imagenet}
\bibinfo{author}{Krizhevsky, A.}, \bibinfo{author}{Sutskever, I.}, \bibinfo{author}{Hinton, G.E.}, \bibinfo{year}{2012}.
\newblock \bibinfo{title}{Imagenet classification with deep convolutional neural networks}.
\newblock \bibinfo{journal}{Advances in neural information processing systems} \bibinfo{volume}{25}.
\bibitem[{LeCun et~al.(1998)LeCun, Bottou, Bengio and Haffner}]{lecun1998gradient}
\bibinfo{author}{LeCun, Y.}, \bibinfo{author}{Bottou, L.}, \bibinfo{author}{Bengio, Y.}, \bibinfo{author}{Haffner, P.}, \bibinfo{year}{1998}.
\newblock \bibinfo{title}{Gradient-based learning applied to document recognition}.
\newblock \bibinfo{journal}{Proceedings of the IEEE} \bibinfo{volume}{86}, \bibinfo{pages}{2278--2324}.
\bibitem[{Li et~al.(2022a)Li, Li, Xiong and Hoi}]{li2022blip}
\bibinfo{author}{Li, J.}, \bibinfo{author}{Li, D.}, \bibinfo{author}{Xiong, C.}, \bibinfo{author}{Hoi, S.}, \bibinfo{year}{2022}a.
\newblock \bibinfo{title}{Blip: Bootstrapping language-image pre-training for unified vision-language understanding and generation}, in: \bibinfo{booktitle}{International conference on machine learning}, \bibinfo{organization}{PMLR}. pp. \bibinfo{pages}{12888--12900}.
\bibitem[{Li et~al.(2021)Li, Selvaraju, Gotmare, Joty, Xiong and Hoi}]{li2021align}
\bibinfo{author}{Li, J.}, \bibinfo{author}{Selvaraju, R.}, \bibinfo{author}{Gotmare, A.}, \bibinfo{author}{Joty, S.}, \bibinfo{author}{Xiong, C.}, \bibinfo{author}{Hoi, S.C.H.}, \bibinfo{year}{2021}.
\newblock \bibinfo{title}{Align before fuse: Vision and language representation learning with momentum distillation}.
\newblock \bibinfo{journal}{Advances in neural information processing systems} \bibinfo{volume}{34}, \bibinfo{pages}{9694--9705}.
\bibitem[{Li et~al.(2022b)Li, Zhang, Zhang, Yang, Li, Zhong, Wang, Yuan, Zhang, Hwang et~al.}]{li2022grounded}
\bibinfo{author}{Li, L.H.}, \bibinfo{author}{Zhang, P.}, \bibinfo{author}{Zhang, H.}, \bibinfo{author}{Yang, J.}, \bibinfo{author}{Li, C.}, \bibinfo{author}{Zhong, Y.}, \bibinfo{author}{Wang, L.}, \bibinfo{author}{Yuan, L.}, \bibinfo{author}{Zhang, L.}, \bibinfo{author}{Hwang, J.N.}, et~al., \bibinfo{year}{2022}b.
\newblock \bibinfo{title}{Grounded language-image pre-training}, in: \bibinfo{booktitle}{Proceedings of the IEEE/CVF Conference on Computer Vision and Pattern Recognition}, pp. \bibinfo{pages}{10965--10975}.
\bibitem[{Li et~al.(2022c)Li, Huang, Zhu, Tang, Li, Zhou and Lu}]{li2022ordinalclip}
\bibinfo{author}{Li, W.}, \bibinfo{author}{Huang, X.}, \bibinfo{author}{Zhu, Z.}, \bibinfo{author}{Tang, Y.}, \bibinfo{author}{Li, X.}, \bibinfo{author}{Zhou, J.}, \bibinfo{author}{Lu, J.}, \bibinfo{year}{2022}c.
\newblock \bibinfo{title}{Ordinalclip: Learning rank prompts for language-guided ordinal regression}.
\newblock \bibinfo{journal}{Advances in Neural Information Processing Systems} \bibinfo{volume}{35}, \bibinfo{pages}{35313--35325}.
\bibitem[{Liang et~al.(2023)Liang, Xie, Zou, Ye, Xu and Bai}]{liang2023crowdclip}
\bibinfo{author}{Liang, D.}, \bibinfo{author}{Xie, J.}, \bibinfo{author}{Zou, Z.}, \bibinfo{author}{Ye, X.}, \bibinfo{author}{Xu, W.}, \bibinfo{author}{Bai, X.}, \bibinfo{year}{2023}.
\newblock \bibinfo{title}{Crowdclip: Unsupervised crowd counting via vision-language model}, in: \bibinfo{booktitle}{Proceedings of the IEEE/CVF Conference on Computer Vision and Pattern Recognition}, pp. \bibinfo{pages}{2893--2903}.
\bibitem[{Liu et~al.(2024)Liu, Qi, Li, Hassanpour, Wang, Plataniotis and Yu}]{liu2024test}
\bibinfo{author}{Liu, H.}, \bibinfo{author}{Qi, J.}, \bibinfo{author}{Li, Z.}, \bibinfo{author}{Hassanpour, M.}, \bibinfo{author}{Wang, Y.}, \bibinfo{author}{Plataniotis, K.N.}, \bibinfo{author}{Yu, Y.}, \bibinfo{year}{2024}.
\newblock \bibinfo{title}{Test-time personalization with meta prompt for gaze estimation}, in: \bibinfo{booktitle}{Proceedings of the AAAI Conference on Artificial Intelligence}, pp. \bibinfo{pages}{3621--3629}.
\bibitem[{Liu and Lu(2024)}]{liu2024uvagaze}
\bibinfo{author}{Liu, R.}, \bibinfo{author}{Lu, F.}, \bibinfo{year}{2024}.
\newblock \bibinfo{title}{Uvagaze: Unsupervised 1-to-2 views adaptation for gaze estimation}, in: \bibinfo{booktitle}{Proceedings of the AAAI Conference on Artificial Intelligence}, pp. \bibinfo{pages}{3693--3701}.
\bibitem[{Mahmud et~al.(2024)Mahmud, Hungler and Etemad}]{mahmud2024multistream}
\bibinfo{author}{Mahmud, Z.}, \bibinfo{author}{Hungler, P.}, \bibinfo{author}{Etemad, A.}, \bibinfo{year}{2024}.
\newblock \bibinfo{title}{Multistream gaze estimation with anatomical eye region isolation by synthetic to real transfer learning}.
\newblock \bibinfo{journal}{IEEE Transactions on Artificial Intelligence} .
\bibitem[{Park et~al.(2019)Park, Mello, Molchanov, Iqbal, Hilliges and Kautz}]{park2019few}
\bibinfo{author}{Park, S.}, \bibinfo{author}{Mello, S.D.}, \bibinfo{author}{Molchanov, P.}, \bibinfo{author}{Iqbal, U.}, \bibinfo{author}{Hilliges, O.}, \bibinfo{author}{Kautz, J.}, \bibinfo{year}{2019}.
\newblock \bibinfo{title}{Few-shot adaptive gaze estimation}, in: \bibinfo{booktitle}{Proceedings of the IEEE/CVF international conference on computer vision}, pp. \bibinfo{pages}{9368--9377}.
\bibitem[{Park et~al.(2018)Park, Spurr and Hilliges}]{park2018deep}
\bibinfo{author}{Park, S.}, \bibinfo{author}{Spurr, A.}, \bibinfo{author}{Hilliges, O.}, \bibinfo{year}{2018}.
\newblock \bibinfo{title}{Deep pictorial gaze estimation}, in: \bibinfo{booktitle}{Proceedings of the European conference on computer vision (ECCV)}, pp. \bibinfo{pages}{721--738}.
\bibitem[{Radford et~al.(2021)Radford, Kim, Hallacy, Ramesh, Goh, Agarwal, Sastry, Askell, Mishkin, Clark et~al.}]{radford2021learning}
\bibinfo{author}{Radford, A.}, \bibinfo{author}{Kim, J.W.}, \bibinfo{author}{Hallacy, C.}, \bibinfo{author}{Ramesh, A.}, \bibinfo{author}{Goh, G.}, \bibinfo{author}{Agarwal, S.}, \bibinfo{author}{Sastry, G.}, \bibinfo{author}{Askell, A.}, \bibinfo{author}{Mishkin, P.}, \bibinfo{author}{Clark, J.}, et~al., \bibinfo{year}{2021}.
\newblock \bibinfo{title}{Learning transferable visual models from natural language supervision}, in: \bibinfo{booktitle}{International conference on machine learning}, \bibinfo{organization}{PMLR}. pp. \bibinfo{pages}{8748--8763}.
\bibitem[{Radford et~al.(2018)Radford, Narasimhan, Salimans, Sutskever et~al.}]{radford2018improving}
\bibinfo{author}{Radford, A.}, \bibinfo{author}{Narasimhan, K.}, \bibinfo{author}{Salimans, T.}, \bibinfo{author}{Sutskever, I.}, et~al., \bibinfo{year}{2018}.
\newblock \bibinfo{title}{Improving language understanding by generative pre-training} .
\bibitem[{Radford et~al.(2019)Radford, Wu, Child, Luan, Amodei, Sutskever et~al.}]{radford2019language}
\bibinfo{author}{Radford, A.}, \bibinfo{author}{Wu, J.}, \bibinfo{author}{Child, R.}, \bibinfo{author}{Luan, D.}, \bibinfo{author}{Amodei, D.}, \bibinfo{author}{Sutskever, I.}, et~al., \bibinfo{year}{2019}.
\newblock \bibinfo{title}{Language models are unsupervised multitask learners}.
\newblock \bibinfo{journal}{OpenAI blog} \bibinfo{volume}{1}, \bibinfo{pages}{9}.
\bibitem[{Raffel et~al.(2020)Raffel, Shazeer, Roberts, Lee, Narang, Matena, Zhou, Li and Liu}]{raffel2020exploring}
\bibinfo{author}{Raffel, C.}, \bibinfo{author}{Shazeer, N.}, \bibinfo{author}{Roberts, A.}, \bibinfo{author}{Lee, K.}, \bibinfo{author}{Narang, S.}, \bibinfo{author}{Matena, M.}, \bibinfo{author}{Zhou, Y.}, \bibinfo{author}{Li, W.}, \bibinfo{author}{Liu, P.J.}, \bibinfo{year}{2020}.
\newblock \bibinfo{title}{Exploring the limits of transfer learning with a unified text-to-text transformer}.
\newblock \bibinfo{journal}{Journal of machine learning research} \bibinfo{volume}{21}, \bibinfo{pages}{1--67}.
\bibitem[{Ramesh et~al.(2022)Ramesh, Dhariwal, Nichol, Chu and Chen}]{ramesh2022hierarchical}
\bibinfo{author}{Ramesh, A.}, \bibinfo{author}{Dhariwal, P.}, \bibinfo{author}{Nichol, A.}, \bibinfo{author}{Chu, C.}, \bibinfo{author}{Chen, M.}, \bibinfo{year}{2022}.
\newblock \bibinfo{title}{Hierarchical text-conditional image generation with clip latents}.
\newblock \bibinfo{journal}{arXiv preprint arXiv:2204.06125} \bibinfo{volume}{1}, \bibinfo{pages}{3}.
\bibitem[{Ramesh et~al.(2021)Ramesh, Pavlov, Goh, Gray, Voss, Radford, Chen and Sutskever}]{ramesh2021zero}
\bibinfo{author}{Ramesh, A.}, \bibinfo{author}{Pavlov, M.}, \bibinfo{author}{Goh, G.}, \bibinfo{author}{Gray, S.}, \bibinfo{author}{Voss, C.}, \bibinfo{author}{Radford, A.}, \bibinfo{author}{Chen, M.}, \bibinfo{author}{Sutskever, I.}, \bibinfo{year}{2021}.
\newblock \bibinfo{title}{Zero-shot text-to-image generation}, in: \bibinfo{booktitle}{International conference on machine learning}, \bibinfo{organization}{Pmlr}. pp. \bibinfo{pages}{8821--8831}.
\bibitem[{Shen et~al.(2021)Shen, Li, Tan, Bansal, Rohrbach, Chang, Yao and Keutzer}]{shen2021much}
\bibinfo{author}{Shen, S.}, \bibinfo{author}{Li, L.H.}, \bibinfo{author}{Tan, H.}, \bibinfo{author}{Bansal, M.}, \bibinfo{author}{Rohrbach, A.}, \bibinfo{author}{Chang, K.W.}, \bibinfo{author}{Yao, Z.}, \bibinfo{author}{Keutzer, K.}, \bibinfo{year}{2021}.
\newblock \bibinfo{title}{How much can clip benefit vision-and-language tasks?}
\newblock \bibinfo{journal}{arXiv preprint arXiv:2107.06383} .
\bibitem[{Shi et~al.(2022)Shi, Hayat, Wu and Cai}]{shi2022proposalclip}
\bibinfo{author}{Shi, H.}, \bibinfo{author}{Hayat, M.}, \bibinfo{author}{Wu, Y.}, \bibinfo{author}{Cai, J.}, \bibinfo{year}{2022}.
\newblock \bibinfo{title}{Proposalclip: Unsupervised open-category object proposal generation via exploiting clip cues}, in: \bibinfo{booktitle}{Proceedings of the IEEE/CVF Conference on Computer Vision and Pattern Recognition}, pp. \bibinfo{pages}{9611--9620}.
\bibitem[{Simonyan and Zisserman(2014)}]{simonyan2014very}
\bibinfo{author}{Simonyan, K.}, \bibinfo{author}{Zisserman, A.}, \bibinfo{year}{2014}.
\newblock \bibinfo{title}{Very deep convolutional networks for large-scale image recognition}.
\newblock \bibinfo{journal}{arXiv preprint arXiv:1409.1556} .
\bibitem[{Strazdas et~al.(2022)Strazdas, Hintz, Khalifa, Abdelrahman, Hempel and Al-Hamadi}]{strazdas2022robot}
\bibinfo{author}{Strazdas, D.}, \bibinfo{author}{Hintz, J.}, \bibinfo{author}{Khalifa, A.}, \bibinfo{author}{Abdelrahman, A.A.}, \bibinfo{author}{Hempel, T.}, \bibinfo{author}{Al-Hamadi, A.}, \bibinfo{year}{2022}.
\newblock \bibinfo{title}{Robot system assistant (rosa): Towards intuitive multi-modal and multi-device human-robot interaction}.
\newblock \bibinfo{journal}{Sensors} \bibinfo{volume}{22}, \bibinfo{pages}{923}.
\bibitem[{Vaswani et~al.(2017)Vaswani, Shazeer, Parmar, Uszkoreit, Jones, Gomez, Kaiser and Polosukhin}]{vaswani2017attention}
\bibinfo{author}{Vaswani, A.}, \bibinfo{author}{Shazeer, N.}, \bibinfo{author}{Parmar, N.}, \bibinfo{author}{Uszkoreit, J.}, \bibinfo{author}{Jones, L.}, \bibinfo{author}{Gomez, A.N.}, \bibinfo{author}{Kaiser, {\L}.}, \bibinfo{author}{Polosukhin, I.}, \bibinfo{year}{2017}.
\newblock \bibinfo{title}{Attention is all you need}.
\newblock \bibinfo{journal}{Advances in neural information processing systems} \bibinfo{volume}{30}.
\bibitem[{Wang et~al.(2019)Wang, Zhao, Su and Ji}]{wang2019generalizing}
\bibinfo{author}{Wang, K.}, \bibinfo{author}{Zhao, R.}, \bibinfo{author}{Su, H.}, \bibinfo{author}{Ji, Q.}, \bibinfo{year}{2019}.
\newblock \bibinfo{title}{Generalizing eye tracking with bayesian adversarial learning}, in: \bibinfo{booktitle}{Proceedings of the IEEE/CVF conference on computer vision and pattern recognition}, pp. \bibinfo{pages}{11907--11916}.
\bibitem[{Wang and Shen(2017)}]{wang2017deep}
\bibinfo{author}{Wang, W.}, \bibinfo{author}{Shen, J.}, \bibinfo{year}{2017}.
\newblock \bibinfo{title}{Deep visual attention prediction}.
\newblock \bibinfo{journal}{IEEE Transactions on Image Processing} \bibinfo{volume}{27}, \bibinfo{pages}{2368--2378}.
\bibitem[{Wang et~al.(2023)Wang, Shi, De~Mello, Chang and Zhang}]{wang2023investigation}
\bibinfo{author}{Wang, Y.}, \bibinfo{author}{Shi, X.}, \bibinfo{author}{De~Mello, S.}, \bibinfo{author}{Chang, H.J.}, \bibinfo{author}{Zhang, X.}, \bibinfo{year}{2023}.
\newblock \bibinfo{title}{Investigation of architectures and receptive fields for appearance-based gaze estimation}.
\newblock \bibinfo{journal}{arXiv preprint arXiv:2308.09593} .
\bibitem[{Xu et~al.(2022)Xu, Zhang, Wei, Lin, Cao, Hu and Bai}]{xu2022simple}
\bibinfo{author}{Xu, M.}, \bibinfo{author}{Zhang, Z.}, \bibinfo{author}{Wei, F.}, \bibinfo{author}{Lin, Y.}, \bibinfo{author}{Cao, Y.}, \bibinfo{author}{Hu, H.}, \bibinfo{author}{Bai, X.}, \bibinfo{year}{2022}.
\newblock \bibinfo{title}{A simple baseline for open-vocabulary semantic segmentation with pre-trained vision-language model}, in: \bibinfo{booktitle}{European Conference on Computer Vision}, \bibinfo{organization}{Springer}. pp. \bibinfo{pages}{736--753}.
\bibitem[{Xu et~al.(2018)Xu, Dong, Wu, Sun, Shi, Yu and Gao}]{xu2018gaze}
\bibinfo{author}{Xu, Y.}, \bibinfo{author}{Dong, Y.}, \bibinfo{author}{Wu, J.}, \bibinfo{author}{Sun, Z.}, \bibinfo{author}{Shi, Z.}, \bibinfo{author}{Yu, J.}, \bibinfo{author}{Gao, S.}, \bibinfo{year}{2018}.
\newblock \bibinfo{title}{Gaze prediction in dynamic 360 immersive videos}, in: \bibinfo{booktitle}{proceedings of the IEEE Conference on Computer Vision and Pattern Recognition}, pp. \bibinfo{pages}{5333--5342}.
\bibitem[{Yin et~al.(2024a)Yin, Wang, Dai and Wu}]{yin2024nerf}
\bibinfo{author}{Yin, P.}, \bibinfo{author}{Wang, J.}, \bibinfo{author}{Dai, J.}, \bibinfo{author}{Wu, X.}, \bibinfo{year}{2024}a.
\newblock \bibinfo{title}{Nerf-gaze: A head-eye redirection parametric model for gaze estimation}, in: \bibinfo{booktitle}{ICASSP 2024-2024 IEEE International Conference on Acoustics, Speech and Signal Processing (ICASSP)}, \bibinfo{organization}{IEEE}. pp. \bibinfo{pages}{2760--2764}.
\bibitem[{Yin et~al.(2024b)Yin, Zeng, Wang and Xie}]{yin2024clip}
\bibinfo{author}{Yin, P.}, \bibinfo{author}{Zeng, G.}, \bibinfo{author}{Wang, J.}, \bibinfo{author}{Xie, D.}, \bibinfo{year}{2024}b.
\newblock \bibinfo{title}{Clip-gaze: towards general gaze estimation via visual-linguistic model}, in: \bibinfo{booktitle}{Proceedings of the AAAI Conference on Artificial Intelligence}, pp. \bibinfo{pages}{6729--6737}.
\bibitem[{Yoon et~al.(2019)Yoon, Baek, Truong and Park}]{yoon2019driver}
\bibinfo{author}{Yoon, H.S.}, \bibinfo{author}{Baek, N.R.}, \bibinfo{author}{Truong, N.Q.}, \bibinfo{author}{Park, K.R.}, \bibinfo{year}{2019}.
\newblock \bibinfo{title}{Driver gaze detection based on deep residual networks using the combined single image of dual near-infrared cameras}.
\newblock \bibinfo{journal}{IEEE Access} \bibinfo{volume}{7}, \bibinfo{pages}{93448--93461}.
\bibitem[{Yu and Koltun(2015)}]{yu2015multi}
\bibinfo{author}{Yu, F.}, \bibinfo{author}{Koltun, V.}, \bibinfo{year}{2015}.
\newblock \bibinfo{title}{Multi-scale context aggregation by dilated convolutions}.
\newblock \bibinfo{journal}{arXiv preprint arXiv:1511.07122} .
\bibitem[{Yu et~al.(2022)Yu, Wang, Vasudevan, Yeung, Seyedhosseini and Wu}]{yu2022coca}
\bibinfo{author}{Yu, J.}, \bibinfo{author}{Wang, Z.}, \bibinfo{author}{Vasudevan, V.}, \bibinfo{author}{Yeung, L.}, \bibinfo{author}{Seyedhosseini, M.}, \bibinfo{author}{Wu, Y.}, \bibinfo{year}{2022}.
\newblock \bibinfo{title}{Coca: Contrastive captioners are image-text foundation models}.
\newblock \bibinfo{journal}{arXiv preprint arXiv:2205.01917} .
\bibitem[{Yu et~al.(2023)Yu, Liu, Hua, Jiang, Ren and Bai}]{yu2023turning}
\bibinfo{author}{Yu, W.}, \bibinfo{author}{Liu, Y.}, \bibinfo{author}{Hua, W.}, \bibinfo{author}{Jiang, D.}, \bibinfo{author}{Ren, B.}, \bibinfo{author}{Bai, X.}, \bibinfo{year}{2023}.
\newblock \bibinfo{title}{Turning a clip model into a scene text detector}, in: \bibinfo{booktitle}{Proceedings of the IEEE/CVF Conference on Computer Vision and Pattern Recognition}, pp. \bibinfo{pages}{6978--6988}.
\bibitem[{Zhang et~al.(2022a)Zhang, Guo, Zhang, Li, Miao, Cui, Qiao, Gao and Li}]{zhang2022pointclip}
\bibinfo{author}{Zhang, R.}, \bibinfo{author}{Guo, Z.}, \bibinfo{author}{Zhang, W.}, \bibinfo{author}{Li, K.}, \bibinfo{author}{Miao, X.}, \bibinfo{author}{Cui, B.}, \bibinfo{author}{Qiao, Y.}, \bibinfo{author}{Gao, P.}, \bibinfo{author}{Li, H.}, \bibinfo{year}{2022}a.
\newblock \bibinfo{title}{Pointclip: Point cloud understanding by clip}, in: \bibinfo{booktitle}{Proceedings of the IEEE/CVF conference on computer vision and pattern recognition}, pp. \bibinfo{pages}{8552--8562}.
\bibitem[{Zhang et~al.(2022b)Zhang, Zeng, Guo and Li}]{zhang2022can}
\bibinfo{author}{Zhang, R.}, \bibinfo{author}{Zeng, Z.}, \bibinfo{author}{Guo, Z.}, \bibinfo{author}{Li, Y.}, \bibinfo{year}{2022}b.
\newblock \bibinfo{title}{Can language understand depth?}, in: \bibinfo{booktitle}{Proceedings of the 30th ACM International Conference on Multimedia}, pp. \bibinfo{pages}{6868--6874}.
\bibitem[{Zhang et~al.(2020)Zhang, Park, Beeler, Bradley, Tang and Hilliges}]{zhang2020eth}
\bibinfo{author}{Zhang, X.}, \bibinfo{author}{Park, S.}, \bibinfo{author}{Beeler, T.}, \bibinfo{author}{Bradley, D.}, \bibinfo{author}{Tang, S.}, \bibinfo{author}{Hilliges, O.}, \bibinfo{year}{2020}.
\newblock \bibinfo{title}{Eth-xgaze: A large scale dataset for gaze estimation under extreme head pose and gaze variation}, in: \bibinfo{booktitle}{Computer Vision--ECCV 2020: 16th European Conference, Glasgow, UK, August 23--28, 2020, Proceedings, Part V 16}, \bibinfo{organization}{Springer}. pp. \bibinfo{pages}{365--381}.
\bibitem[{Zhang et~al.(2018)Zhang, Sugano and Bulling}]{zhang2018revisiting}
\bibinfo{author}{Zhang, X.}, \bibinfo{author}{Sugano, Y.}, \bibinfo{author}{Bulling, A.}, \bibinfo{year}{2018}.
\newblock \bibinfo{title}{Revisiting data normalization for appearance-based gaze estimation}, in: \bibinfo{booktitle}{Proceedings of the 2018 ACM symposium on eye tracking research \& applications}, pp. \bibinfo{pages}{1--9}.
\bibitem[{Zhang et~al.(2015)Zhang, Sugano, Fritz and Bulling}]{zhang2015appearance}
\bibinfo{author}{Zhang, X.}, \bibinfo{author}{Sugano, Y.}, \bibinfo{author}{Fritz, M.}, \bibinfo{author}{Bulling, A.}, \bibinfo{year}{2015}.
\newblock \bibinfo{title}{Appearance-based gaze estimation in the wild}, in: \bibinfo{booktitle}{Proceedings of the IEEE conference on computer vision and pattern recognition}, pp. \bibinfo{pages}{4511--4520}.
\bibitem[{Zhang et~al.(2017a)Zhang, Sugano, Fritz and Bulling}]{zhang2017s}
\bibinfo{author}{Zhang, X.}, \bibinfo{author}{Sugano, Y.}, \bibinfo{author}{Fritz, M.}, \bibinfo{author}{Bulling, A.}, \bibinfo{year}{2017}a.
\newblock \bibinfo{title}{It's written all over your face: Full-face appearance-based gaze estimation}, in: \bibinfo{booktitle}{Proceedings of the IEEE conference on computer vision and pattern recognition workshops}, pp. \bibinfo{pages}{51--60}.
\bibitem[{Zhang et~al.(2017b)Zhang, Sugano, Fritz and Bulling}]{zhang2017mpiigaze}
\bibinfo{author}{Zhang, X.}, \bibinfo{author}{Sugano, Y.}, \bibinfo{author}{Fritz, M.}, \bibinfo{author}{Bulling, A.}, \bibinfo{year}{2017}b.
\newblock \bibinfo{title}{Mpiigaze: Real-world dataset and deep appearance-based gaze estimation}.
\newblock \bibinfo{journal}{IEEE transactions on pattern analysis and machine intelligence} \bibinfo{volume}{41}, \bibinfo{pages}{162--175}.
\bibitem[{Zhou et~al.(2022)Zhou, Yang, Loy and Liu}]{zhou2022conditional}
\bibinfo{author}{Zhou, K.}, \bibinfo{author}{Yang, J.}, \bibinfo{author}{Loy, C.C.}, \bibinfo{author}{Liu, Z.}, \bibinfo{year}{2022}.
\newblock \bibinfo{title}{Conditional prompt learning for vision-language models}, in: \bibinfo{booktitle}{Proceedings of the IEEE/CVF conference on computer vision and pattern recognition}, pp. \bibinfo{pages}{16816--16825}.

\end{thebibliography}

\end{document}